\documentclass{article}

\usepackage{microtype}
\usepackage{graphicx}
\usepackage{subfigure}
\usepackage{tabularx}
\usepackage{booktabs} 
\usepackage{multirow}
\usepackage[table]{xcolor}
\usepackage{color}
\usepackage{tablefootnote}
\definecolor{cgreen}{RGB}{79, 121, 66}
\definecolor{cblue}{RGB}{0,91,150}

\usepackage{chngcntr}

\usepackage{hyperref}

\usepackage[accepted]{icml2025}

\usepackage{amsmath}
\usepackage{amssymb}
\usepackage{mathtools}
\usepackage{amsthm}

\usepackage[capitalize,noabbrev]{cleveref}
\usepackage{tikz}

\theoremstyle{plain}

\theoremstyle{definition}

\theoremstyle{remark}

\usepackage[nolist]{acronym}

\usepackage{xspace}

\usepackage{amssymb}
\usepackage{xspace}

\newcommand{\eg}{e.\nolinebreak[4]\hspace{0.01em}\nolinebreak[4]g.\@\xspace}

\icmltitlerunning{DUNIA: Pixel-Sized Embeddings via Cross-Modal Alignment for Earth Observation Applications}

\begin{document}

\twocolumn[
\icmltitle{DUNIA: Pixel-Sized Embeddings via Cross-Modal Alignment for Earth Observation Applications}

\icmlsetsymbol{equal}{*}

\begin{icmlauthorlist}
\icmlauthor{Ibrahim Fayad}{lsce,kayrros}
\icmlauthor{Max Zimmer}{zib}
\icmlauthor{Martin Schwartz}{lsce}
\icmlauthor{Fabian Gieseke}{munster}
\icmlauthor{Philippe Ciais}{lsce}
\icmlauthor{Gabriel Belouze}{lsce}
\icmlauthor{Sarah Brood}{ens}
\icmlauthor{Aurelien De Truchis}{kayrros}
\icmlauthor{Alexandre d'Aspremont}{ens,kayrros}
\end{icmlauthorlist}

\icmlaffiliation{lsce}{Laboratoire des Sciences du Climat et de l’Environnement,
LSCE/IPSL, France}
\icmlaffiliation{kayrros}{Kayrros SAS, Paris 75009, France}
\icmlaffiliation{zib}{Department for AI in Society,
Science, and Technology, Zuse Institute Berlin, Germany}
\icmlaffiliation{munster}{Department of Information Systems, University of Münster, Germany}
\icmlaffiliation{ens}{Department of Computer Science, CNRS, INRIA \& École Normale Supérieure, Paris 75230, France}

\icmlcorrespondingauthor{Ibrahim Fayad}{ifayad@lsce.ipsl.fr}

\icmlkeywords{Machine Learning, ICML}

\vskip 0.3in
]

\printAffiliationsAndNotice

\begin{abstract}
   Significant efforts have been directed towards adapting self-supervised multimodal learning for Earth observation applications. However, most current methods produce coarse patch-sized embeddings, limiting their effectiveness and integration with other modalities like LiDAR. To close this gap, we present DUNIA, an approach to learn pixel-sized embeddings through cross-modal alignment between images and full-waveform LiDAR data. As the model is trained in a contrastive manner, the embeddings can be directly leveraged in the context of a variety of environmental monitoring tasks in a zero-shot setting. In our experiments, we demonstrate the effectiveness of the embeddings for seven such tasks: canopy height mapping, fractional canopy cover, land cover mapping, tree species identification, plant area index, crop type classification, and per-pixel waveform-based vertical structure mapping. The results show that the embeddings, along with zero-shot classifiers, often outperform specialized supervised models, even in low-data regimes. In the fine-tuning setting, we show strong performances near or better than the state-of-the-art on five out of six tasks. 
\end{abstract}

\begin{figure}[ht]
\includegraphics[width=\linewidth]{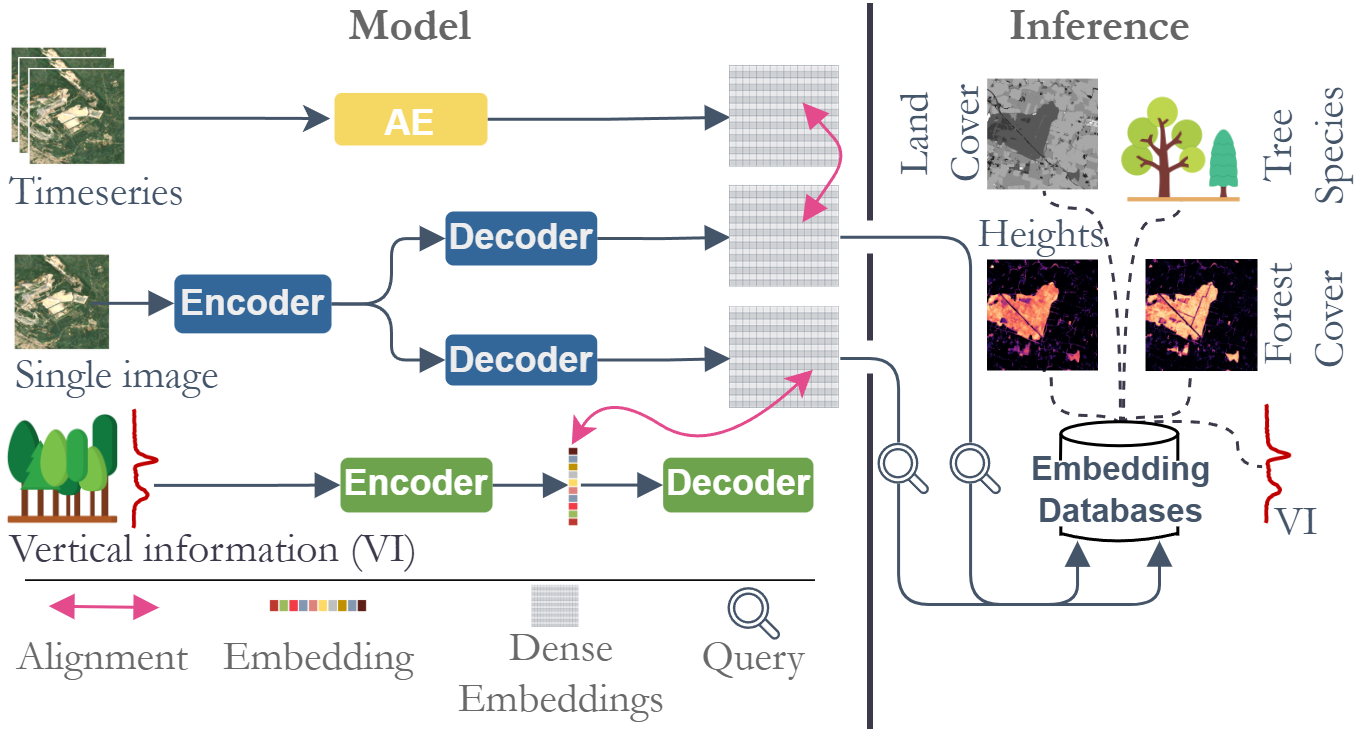}
\vspace{-\baselineskip}
\caption{DUNIA’s alignment strategy leverages temporal and vertical information at the pixel level, enabling the model to develop a cross-modal understanding from a single satellite image. This approach serves as a foundation for cross-modal retrieval tasks across diverse EO applications. The pre-trained model is represented in navy blue, the multi-temporal autoencoder (AE) in yellow, and the waveform AE in green.}
\vspace{-\baselineskip}
\label{fig:overview}
\end{figure}

\vspace{-1\baselineskip}
\section{Introduction}
With the rapid expansion of Earth Observation~(EO) satellite missions, deep learning has emerged as a powerful solution for key EO applications, ranging from monitoring natural resources \cite{Chen_etal_2023, Fayad_etal_2024, Li_etal_2023a, Liu_etal_2022, liu2023overlooked, Tolan_etal_2024, Lang_etal_2023, Pauls_etal_2024, schwartz2023forms, pauls2025capturing} and assessing environmental impacts \cite{Dalagnol_etal_2023, Wagner_etal_2023}, to climate monitoring and forecast \cite{Andrychowicz_etal_2023, Schultz_etal_2021}, as well as agricultural and land resource management \cite{Ienco_etal_2019, Kussul_etal_2017, Zhang_etal_2019}.

While research of new deep learning architectures and upcoming remote satellite missions is expected to enhance existing EO applications, 
supervised deep learning approaches have key limitations. First, these models depend on labeled data, often requiring large annotated datasets. For example, the estimation of above-ground biomass~(AGB) still heavily relies on traditional methods due to limited ground-based measurements and the need to use other variables as proxies~\cite{Morin_etal_2023, schwartz2023forms}. 

Second, EO models are often tailored to specific tasks, limiting their adaptability and reusability, even when using similar input data sources. For instance, a land cover classification model cannot easily predict canopy height due to differences in input-output relationships, as the network may prioritize learning features that are highly discriminative for a given task but not the other. 

Third, EO models often use optical or radar imagery to predict single-valued targets like canopy height or land cover classes. However, they struggle with more complex outputs like the full vertical structure of vegetation.\footnote{Capturing such vertical structures is essential for understanding biomass allocation and species diversity. Predicting vertical structure from EO imagery is challenging due to its complexity and limited sensor data, requiring cross-modal understanding over direct prediction \cite{Tan_etal_2024}.}

Foundation models (FMs) offer a promising solution to the aforementioned limitations by leveraging self-supervised pre-training on vast amounts of unlabeled data \cite{Brown_etal_2020,Devlin_2018,Kirillov_etal_2023,Minderer_etal_2022,Radford_etal_2021}. For EO applications, such models are being developed to process diverse data types, including time series processing \cite{Yuan_etal_2022}, adaptation to different input satellite sensors \cite{Xiong_etal_2024} or varying Ground Sample Distances (GSD) \cite{Reed_etal_2023}, and multi-modal data fusion \cite{Astruc_etal_2025, astruc2024anysat, Fuller_etal_2024}.
Most of current foundation models primarily address the adaptability issue, but not label scarcity. In fact, most EO-focused FMs are pre-trained masked auto-encoders (MAEs), which require extensive fine-tuning \cite{Lehner_etal_2024,Singh_etal_2023}. Moreover, existing FMs, even multi-modal ones, only leverage image-based modalities. 
As such, they excel in EO tasks requiring horizontal structural understanding (e.g., land cover mapping, tree species identification), but struggle with EO applications requiring vertical structure understanding (e.g., canopy height estimation).
Additionally, their patch-based approach limits direct pixel-level prediction capabilities. This hinders the integration of LiDAR data, such as full-waveform LiDAR data available as labels at resolutions higher than the patch scale. 

In this work, we propose a \emph{Dense Unsupervised Nature Interpretation Algorithm} (DUNIA) that learns to generate pixel-level embeddings by aligning vertical (from a space-borne full waveform LiDAR) and horizontal structure information with satellite imagery through contrastive learning. This dual alignment strategy enables understanding of both vertical and horizontal structures, supporting diverse EO applications with minimal, if any, training. DUNIA achieves strong retrieval performance for several tasks and excels in forest structure benchmarks. In our experiments, we demonstrate the effectiveness of the resulting embeddings for seven key EO applications and show that zero-shot classification operating on these embeddings can outperform specialized supervised EO models in several cases. A simplified version of our approach is presented in ~\cref{fig:overview}.

\section{Background \& Contribution}
We sketch the related works along with our contributions. 

\subsection{Contrastive Learning}
Contrastive learning trains networks to produce similar embeddings for semantically related inputs and dissimilar embeddings for unrelated ones. Related inputs may include different augmentations of the same sample or paired examples from different modalities. In this context, one of the main challenges is to avoid model collapse, where model outputs become constant. This is typically addressed by: 1) architectural designs or 2) specialized objective functions. 

Prominent examples of architectural designs are BYOL~\cite{Grill_etal_2020} and SimSiam~\cite{Chen_etal_2021}, which respectively use asymmetry to prevent collapse, BYOL employing a momentum encoder and SimSiam using a stop-gradient operation with a predictor network. 
Specialized objective functions can be categorized into pair-based contrastive losses, clustering losses, and negative-pair-free losses. Here, pair-based losses, such as those presented in SimCLR~\cite{Chen_etal_2020} and MoCo~\cite{He_etal_2020}, align positive pairs and separate negative pairs, but assume unique positive pairs within a mini-batch, which is not always true for EO data, especially at high resolutions. Clustering methods like SeLa~\cite{Asano_etal_2019} and SwAV~\cite{Caron_etal_2020} relax this constraint by optimizing a contrastive objective over cluster assignments instead. Negative-pair-free methods, such as Barlow Twins~\cite{Zbontar_etal_2021} and VICReg~\cite{bardes2021vicreg}, focus on redundancy reduction in the feature dimension. Zero-CL \cite{Zhang_etal_2021} uses whitening transformations on the embeddings to maximize the trace of the cross-covariance matrix, achieving alignment and redundancy reduction without negative pairs.

\subsection{Self-Supervised Learning for EO}
Inspired by masked image modeling (MIM) from vision transformers, several SSL frameworks have adapted MAE-style objectives to the remote sensing domain. These methods mainly focus on reconstructing masked inputs to learn spatial, spectral, and temporal representations. SatMAE ~\cite{cong2022satmae} introduces a spectral-temporal aware masking scheme for Sentinel-2 (S-2) timeseries. SatMAE++ ~\cite{noman2024satmaePP} extends SatMAE by incorporating multi-scale feature extraction. Scale-MAE ~\cite{Reed_etal_2023} embeds spatial resolution information into positional encodings for multiscale imagery. DOFA ~\cite{Xiong_etal_2024} extends masked autoencoding with a dynamic, spectral-aware weighting mechanism, enabling the model to encode diverse sensor modalities through a shared encoder.

Other related works on SSL for EO applications relies on a contrastive objective to capture the relationships within the data. CROMA ~\cite{fuller2023croma} aligns radar and optical modalities using a cross-modal contrastive loss, supported by an auxiliary reconstruction head. DeCUR ~\cite{wang2024decur} explicitly disentangles intra- and inter-modal contrastive signals based on Barlow Twins ~\cite{Zbontar_etal_2021}. AnySat ~\cite{astruc2024anysat} proposed a resolution- and temporal-adaptive framework capable of handling data from various sensors and resolutions, effectively combining multiple EO data sources into a shared embedding space.

\subsection{Multi-Modal Learning} 
Multi-modal SSL approaches can be categorized into three main categories. (1) Modality-agnostic models use a shared network for different modalities~\cite{Carreira_etal_2022,Girdhar_etal_2022,Jaegle_etal_2021}, but are typically limited to one modality at a time during training, restricting cross-modal knowledge sharing \cite{Srivastava_etal_2024}. For EO applications, several models follow this approach. Scale-MAE~\cite{Reed_etal_2023} and DOFA~\cite{Xiong_etal_2024} both process imagery across varying resolutions, the latter also processes different wavelengths using a common encoder. (2) Fusion-encoder models integrate data from multiple modalities through cross-modal attention, effectively combining information from each modality~\cite{Bachmann_etal_2022, Bao_etal_2021, Singh_etal_2022}. OmniSat~\cite{Astruc_etal_2025} and AnySat~\cite{astruc2024anysat} exemplify this by integrating data from diverse EO sources into a unified representation. (3) Finally, multi-encoder approaches use separate encoders for each modality, which are then aligned in a shared latent space. This strategy is highly effective for tasks such as cross-modal retrieval and zero-shot classification across different modalities: image/text~\citep{Radford_etal_2021, Jia_etal_2021, Yuan_etal_2021, Yu_etal_2022}, audio/text~\cite{Guzhov_etal_2022}, and video/text~\citep{Luo_etal_2022, Pei_etal_2023}. 

\subsection{Contribution}
Our work aligns closely with the abovementioned third category, focusing on mono- and cross-modal retrieval within the multi-encoder approach. Unlike prior models that align paired data at the instance level (\eg, matching an image to its textual description), here we perform pixel-level alignment. This is necessary to preserve the spatial resolution required to match with full LiDAR waveform footprints.

Specifically, our approach aligns sparse LiDAR waveforms with high-resolution imagery through contrastive learning to understand the vertical structure (i.e., pixel-waveform alignment) and simultaneously performs pixel-pixel alignment for horizontal structure understanding. DUNIA enables per-pixel vertical and horizontal structure retrieval and surpasses specialized models in forest structure benchmarks. For land cover understanding, it performs competitively in zero-shot and low-shot evaluations. Additionally, pixel embeddings can directly generate waveforms representing forest vertical structures — a task not possible with existing methods.

\section{Approach}
Our objective is to learn two distinct pixel-level embedding spaces corresponding to two EO data modalities — vertical and horizontal — using a single pre-trained model. This enables simultaneous understanding of both horizontal and vertical structures. Vertical information is derived from the Global Ecosystem Dynamics Investigation (GEDI) instrument \cite{Dubayah_etal_2020}, a spaceborne LiDAR that provides sparse measurements of 3D structures via 1D waveform signals. Horizontal information comes from 10 m resolution imagery acquired by the S-1 \& 2 missions.

While our framework can be implemented in various ways, we adopt a modular design centered around a transformer encoder and two independent convolutional decoders. This encoder-decoder architecture enables us to generate pixel-sized embeddings for each input. The two decoders are designed to disentangle horizontal and vertical structure understanding: one focuses on spatial (horizontal) relationships, the other on vertical structure.

For horizontal structure alignment, we use a multi-temporal autoencoder (AE) that encodes a sequence of satellite images and produces temporally informed, per-pixel embeddings. This design achieves two goals: 1) it provides an augmented input to the pre-trained model, and 2) it allows the model to implicitly capture phenological patterns without requiring a time series at inference time.

For vertical structure alignment, we also use an AE, this time encoding GEDI waveforms. The waveform encoder projects the 1D input into a latent space, which is aligned with the corresponding pixel embedding from the pre-trained model. While the decoder is not needed for the alignment itself, it is necessary for waveform generation, ensuring that the latent space remains semantically and structurally meaningful.

\begin{figure*}[ht]
\begin{center}
\includegraphics[width=\textwidth]{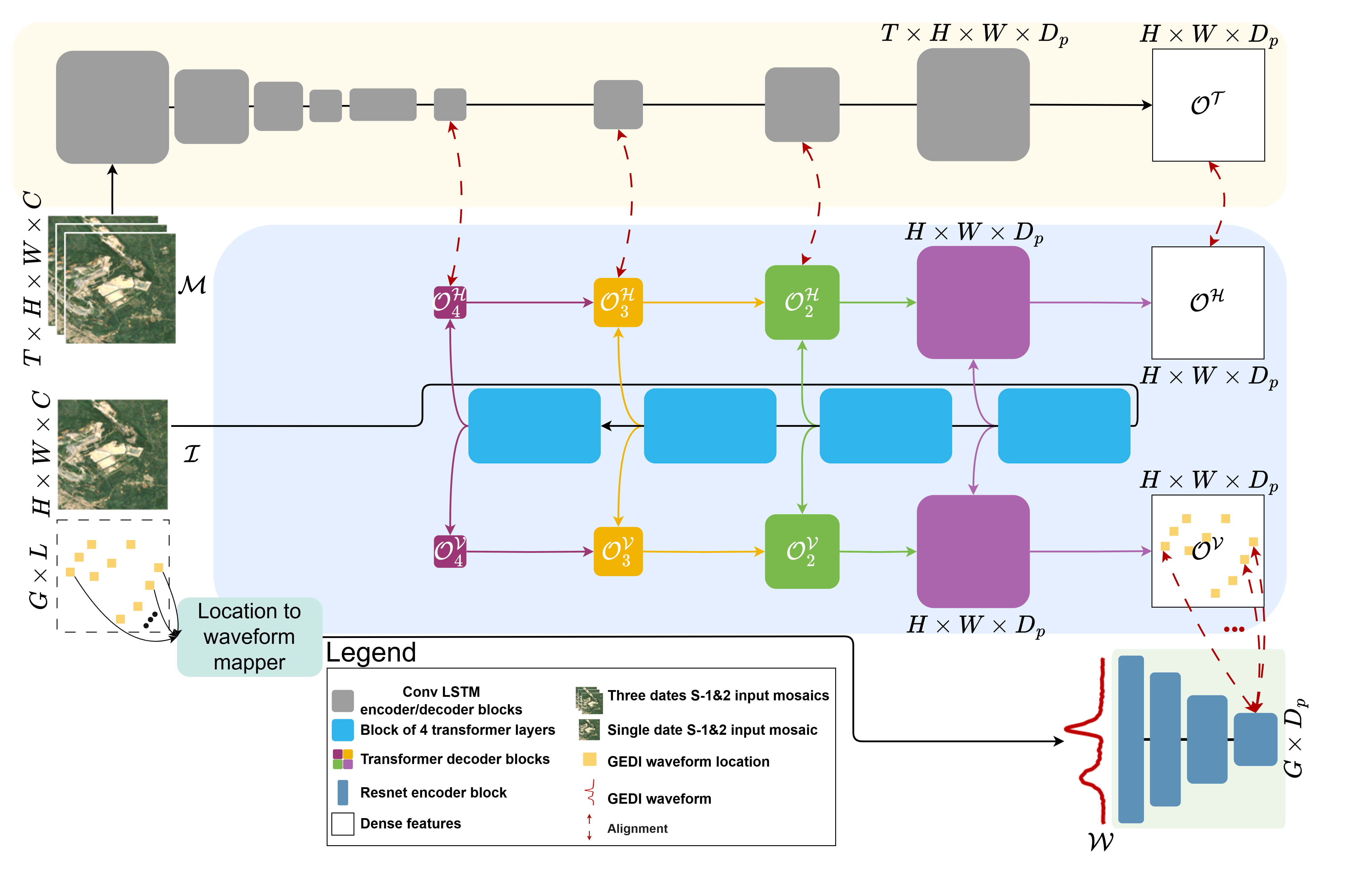}
\end{center}
\vskip-0.3cm
\caption{Simplified architectural overview of the proposed framework. The figure illustrates the key encoder and decoder components, input modalities, and the layers where alignment is performed (red dashed arrows). The pre-trained model is shaded in blue. The multi-temporal image AE, is shaded in yellow. The waveform encoder is shaded in green. Additional components, losses and training objectives are omitted for clarity; see \cref{fig:dunia_full_overview} for full details.}
\label{fig:dunia_overview}
\vskip-0.14cm
\end{figure*}

Figure~\ref{fig:dunia_overview} illustrates the main components of our approach, including the inputs to each model, their architectural building blocks, and the layers used for alignment. Below, we describe the main modules in more detail, followed by our pre-training objectives and the latent diffusion model for waveform generation. Full details are provided in \cref{appendix:implementation_details}.

\subsection{Pre-trained Model}
\label{sec:image-encodec}
The pre-trained model in DUNIA (\cref{fig:dunia_overview}) takes an image $\mathcal{I} \in \mathbb{R}^{H \times W \times C}$ and produces pixel level embeddings in $\mathbb{R}^{H \times W \times D_p}$. $C$ is the number of input channels, $(H,W)$ is the image resolution, and $D_p$ is the target projection dimension. $\mathcal{I}$ is a median composite image generated from S-1 \& 2 observations over several dates. The pre-trained model comprises five building blocks: a patch embedding, a shared encoder, two similar decoders, neighborhood attention layers and several projection heads. Following is a brief description of each of these blocks. 

\paragraph{Patch Embedding.} The 2D composite image $\mathcal{I}$ is split into $ N = HW/P^2$ patches, each of dimension $D$ using a convolutional layer, with $P$ and $D$ the patch size and the patch embedding dimension respectively.

\paragraph{Encoder Architecture.} The encoder consists of 16 standard transformer layers \citep{vaswani}, organized into four Transformer Blocks, of four layers each. Each layer inputs/outputs a sequence of image tokens in $\mathbb{R}^{N \times D}$.

\paragraph{Decoder Architecture.} The two decoders transform the encoder outputs into two sets of pixel-sized embeddings ($\mathcal{O}^\mathcal{V}$, $\mathcal{O}^\mathcal{H} \in \mathbb{R}^{H \times W \times D_p}$) for pixel-waveform and pixel-pixel alignment respectively. They use a hierarchical structure inspired by Ranftl et al. \yrcite{ranftl2021vision} that progressively upsamples features while combining information from different encoder layers to maintain both fine and coarse details. Each decoder block ($d \in \{1,2,3,4\}$, with $d=1$ representing the final decoder layer) produces an image-like feature map of shape $\frac{H}{2^{(d-1)}} \times \frac{W}{2^{(d-1)}} \times D_{p_{d}}$, where \(D_{p_d} = 2^{d-1}D_p\) is the feature dimension out of block $d$. We refer to the output at these stages for a given decoder as $\mathcal{O}_d$. For simplicity, we will henceforth refer to $\mathcal{O}_1$ as $\mathcal{O}$.  

\paragraph{Neighborhood Attention (NA).} To enhance local spatial relationships, we add two NA layers \cite{hassani2023natten} for each decoder block at the highest resolution ($d=1$), where each pixel attends to its surrounding window of size $w$. This complements the global attention from the transformer layers with explicit element relation modeling mechanism.

\paragraph{Projection Head.} For the pixel-pixel alignment objective, we follow standard practice and append a projection head for each output of $\mathcal{O}^\mathcal{H}_d$. As the projection head becomes more specialized towards the training objective \cite{xue2024projectionhead}, its output becomes less generalizable when training and downstream objectives are misaligned, and as such it is discarded after training. For the pixel-waveform alignment $\mathcal{O}^\mathcal{V}$ no projection head is used as the downstream tasks (e.g., waveform generation) align with the training objective.

\subsection{Waveform Model}
\label{sec:waveform_ae}
The waveform AE is based on a VQ-VAE (Vector Quantized Variational Autoencoder) \citep{oordVQVAE} architecture. It consists of three primary components: a waveform encoder (\cref{fig:dunia_overview}), a residual vector quantizer (RVQ), and a waveform decoder. The input to this model are GEDI LiDAR waveforms ($\mathcal{W}$) available only for certain pixel locations in the composite image $\mathcal{I}$.

\paragraph{Waveform Encoder.} The waveform encoder ($\mathcal{E}_w$) uses a ResNet-based architecture to process 1D waveform inputs through multiple stages, progressively reducing their length while increasing feature representation capacity. $\mathcal{E}_w$ transforms a waveform ($\mathcal{W}$) of shape $\mathbb{R}^{L \times 1}$ into a latent representation $z_e \in \mathbb{R}^{L/16 \times 16}$. To align $z_e$ with its corresponding pixel embedding from $\mathcal{O}^\mathcal{V}$ we perform a channel-wise average pooling followed by a linear projection. This process transforms $z_e$ into $\mathcal{O}^\mathcal{W} \in \mathbb{R}^{D_p}$.

\paragraph{Residual Vector Quantizer (RVQ).}\label{paragraph:rvq_layer} For regularization by enforcing discrete representations, we use an RVQ layer with $Q$ quantizers each containing $V$ codebook vectors. The RVQ iteratively quantizes the latent waveform representation $z_e$ through a sequence of quantizers, with each processing the residual from the previous one to produce the final quantized representation $z$.

\paragraph{Waveform Decoder.} The waveform decoder $\mathcal{D}_w$ mirrors the structure of the encoder but operates in reverse order to reconstruct the waveform from its quantized latent representation $z$. In essence, $\mathcal{D}_w(z) = \hat{\mathcal{W}}$. While waveform reconstruction is not necessary for the alignment, reconstructing directly from aligned encoded waveforms ensures that the future generation process operates on a waveform latent space that is already structured based on the shared semantic and structural alignment with the pixel embeddings. This approach is beneficial as we are interested in generating waveforms given pixel inputs.

\subsection{Multitemporal Image Processing Model}\label{sec:multitemproal_ae} The multitemporal image model (\cref{fig:dunia_overview}) is designed to process temporal sequences of median composite images, following an autoencoder (AE) architecture. The input to this model is a sequence of images $\mathcal{M} \in \mathbb{R}^{T \times H \times W \times C}$ covering the same area as the input to the pre-training model. These images come from $T$ S-1 \& 2 acquisition dates that overlap with the dates used to generate the composite image $\mathcal{I}$. This design choice, unlike multi-temporal pre-training models that require time series data during training and inference, or mono-temporal models limited to a single date, is a trade-off that enables the model to capture some phenological traits without needing time series data during inference. Indeed, while image composites such as a median composite may be richer than single-date images, they are still less informative than a full time series, as they only provide a median reflectance value over a given period. Conversely, even this simple form of aggregation preserves significantly more information than a single-date image.

\paragraph{Multi-temporal Image AE.} This AE follows a UNet structure, but replaces conventional convolutional blocks with ConvLSTM \cite{shi2015convlstm} layers to explicitly model temporal correlations. The model takes a multi-temporal input image $\mathcal{M}$ and produces a feature map $\mathcal{X} \in \mathbb{R}^{T \times H \times W \times D_p}$, which is then reconstructed back to a tensor $\mathcal{\hat M}$ of the same shape as the inputs. Each decoder block produces an image-like feature map of shape $\frac{H}{2^{(d-1)}} \times \frac{W}{2^{(d-1)}}$ denoted as $\mathcal{X}_d$ for a respective decoder stage.

\paragraph{Temporal Pooling.} To align the output embeddings from the multi-temporal image AE with those from the pre-training model, we first perform temporal average pooling. This transforms $\mathcal{X}_d \in \mathbb{R}^{T \times \frac{H}{2^{(d-1)}} \times \frac{W}{2^{(d-1)}} \times D_p}$ to $\mathcal{O}^\mathcal{T}_d\in \mathbb{R}^{\frac{H}{2^{(d-1)}} \times \frac{W}{2^{(d-1)}} \times D_p}$, which then passes through a projection head. For the output with the highest resolution (i.e., $\mathcal{O}^\mathcal{T}_1$) we also append two NA layers before the projection head.  For simplicity we refer to $\mathcal{O}^\mathcal{T}_1$ as $\mathcal{O^T}$.

\subsection{Pre-Training Objective}\label{se:pretraining_objective}

DUNIA is pre-trained on pixel-waveform and pixel-pixel alignment, alongside modality reconstruction for the modality-specific AEs. For the alignment task, we rely on two, similarly performing, non-negative-pair contrastive losses, namely VICReg \cite{bardes2021vicreg} and Zero-CL \cite{Zhang_etal_2021}. The selection between either loss functions is determined by the number of available elements within a mini-batch and the requirements of each loss function. VICReg requires large batch sizes to perform well \cite{bardes2021vicreg}, which is a non-issue for the pixel-pixel contrastive objective given the high number of pixels available in each mini-batch of images. However, this is not suitable for the waveforms which are few in number within the mini-batch. On the other hand, Zero-CL performs well even for small batch sizes but faces computational bottlenecks with very large batches. 


\subsubsection{Pixel-Waveform Alignment} Zero-CL replaces the alignment and uniformity terms in negative-pair-based contrastive losses \cite{Arora_etal_2019} with, respectively, an instance-wise contrastive loss ($\mathcal{L}_{\text{Ins}}$) and a feature-wise contrastive loss ($\mathcal{L}_{\text{Fea}}$). The overall loss $\mathcal{L}_{Zero-CL}$ is $\mathcal{L}_{\text{Fea}} + \mathcal{L}_{\text{Ins}}$. $\mathcal{L}_{Zero-CL}$ is applied on \( Z^A \in \mathbb{R}^{G\times D_p} \), the $L_2$ normalized pixel embeddings from $\mathcal{O}^\mathcal{V}$, and \( Z^B \in \mathbb{R}^{G \times D_p} \) their corresponding $L_2$ normalized waveform embeddings from $\mathcal{O}^\mathcal{W}$, where \( G \) is the number of available GEDI samples in a given mini-batch. The formulation for this loss can be found in \cref{appendix:objective_functions}.

\subsubsection{Pixel-Pixel Alignment} VICReg is formulated as three loss terms: variance ($\mathcal{L}_{\text{var}}$), invariance ($\mathcal{L}_{\text{inv}}$), and covariance ($\mathcal{L}_{\text{cov}}$). Let \( Z^\mathcal{H}_{d} \in \mathbb{R}^{M_d\times D_{p_d}} \) represent the $L_2$ normalized pixel embeddings from \( \mathcal{O}^\mathcal{H}_d\), \( M_d = B \times \frac{H}{2^{(d-1)}} \times \frac{W}{2^{(d-1)}} \) with $B$ the mini-batch size, $\frac{H}{2^{(d-1)}}$ and $\frac{W}{2^{(d-1)}}$ the height and width of a decoder's output feature map. \( Z^\mathcal{T}_d \in \mathbb{R}^{M_d \times D_{p_d}} \) is their corresponding $L_2$ normalized pixel embeddings from \( \mathcal{O}^\mathcal{T}_d \). The overall hierarchical pixel-pixel alignment loss is expressed as:
\begin{align}
    \mathcal{L}_\text{VICReg} = \sum_{d=1}^{4} \alpha_v \mathcal{L}_{\text{var}}(Z^\mathcal{H}_{d},Z^\mathcal{T}_{d})\nonumber\\
    + \beta_i\mathcal{L}_{\text{inv}}(Z^\mathcal{H}_{d},Z^\mathcal{T}_{d}) \nonumber\\
    + \gamma_c \mathcal{L}_{\text{cov}}(Z^\mathcal{H}_{d},Z^\mathcal{T}_{d})\label{hiearchical_vicreg}
\end{align}
Following Bardes \yrcite{bardes2021vicreg} we set $\alpha_v$ and $\beta_i$ to 25.0 and $\gamma_c$ to 1.0. VICReg formulation can be found in \cref{appendix:objective_functions}.

\subsubsection{Reconstruction Losses} In addition to the previously defined contrastive losses, our pre-training objective also has three additional reconstruction losses for the modality-specific AEs (i.e., the waveform AE and the multi-temporal image AE) and on the outputs of $\mathcal{O}^{\mathcal{H}}$ for regularization. The reconstruction losses are simply the mean squared error loss (MSE) on the reconstructed waveform $\mathcal{\hat{W}} \in \mathbb{R}^{1 \times L}$ given input waveform $\mathcal{W}$, the reconstructed multi-temporal image $\mathcal{\hat{M}} \in \mathbb{R}^{T \times H \times W \times C}$ given input image $\mathcal{M}$, and the reconstructed mono-temporal image $\mathcal{\hat{I}} \in \mathbb{R}^{H \times W \times C}$ given input image $\mathcal{I}$. The reconstruction term is formulated as:
\begin{align}
    \mathcal{L}_{\text{rec}} = MSE(\mathcal{W},\mathcal{\hat{W}})&+ MSE(\mathcal{M},\mathcal{\hat{M}})\nonumber\\
    &+ MSE(\mathcal{I},\mathcal{\hat{I}})\label{rec_losses}
\end{align}

\subsection{Waveform Generation} \label{sec:diffusion}

To generate a full GEDI waveform given a pixel input, we leverage latent diffusion models (LDMs) \cite{rombach2022sd}. LDMs are generative models that iteratively refine a latent representation, starting from a normally distributed prior in a learned latent space. These models can also be conditioned on auxiliary inputs $y$ to model conditional distributions $p(z|y)$. In our case, the conditioning variable consists of pixel embedding $\mathcal{O}^{\mathcal{V}}_{\phi,\lambda}$ at coordinate $(\phi, \lambda)$, while the latent variable $z \in \mathbb{R}^{L/16\times16}$ represents the embedding of the quantized waveform obtained from the RVQ layer (\cref{paragraph:rvq_layer}). During inference, the waveform generation follows a two-step process. First we sample a representation $z$ given condition $\mathcal{O}^{\mathcal{V}}_{\phi,\lambda}$ using the LDM, and then the frozen $D_w(z)$ yields a waveform.

For a detailed description of the diffusion process, including the denoising objective, noise schedule, and latent variable sampling, we refer the reader to \cref{appendix:diffusion_math}.

\section{Experimental Evaluation}
We provide a detailed experimental evaluation showing that DUNIA is able to achieve high performance across a variety of tasks in zero-shot and fine-tuned settings, including land cover classification, crop mapping, and vertical forest structure analysis (e.g., canopy cover, height and GEDI waveform retrieval). We leverage diverse datasets to inform the model on horizontal structures -- S-1 \& 2, and vertical structures with GEDI waveforms.

The following section contains a summary of the experimental setup, with full details in the appendix. Model configuration and optimisation can be found in Appendix \ref{appendix:model_config_and_optim}. For clarity, during the pre-training phase, we set the input image size to $64\times64$ pixels, with 14 channels from stacked S-1 \& 2 image composites. The embedding dimension is set to 64 (i.e., $D_p=64$). During inference, the input image can be of any size, but we inferred on $256\times256$ pixel images. Our code is available at \url{github.com/AI4Forest/DUNIA}.

\subsection{Experimental Setup}
For our evaluation, we resort to various datasets and tasks. The datasets and experimental details are described next.

\subsubsection{Datasets}

\paragraph{Pre-training Datasets.} \label{seq:pretraining_datasets}

We used S-2 Level-2A surface reflectance data from Google Earth Engine, including 10 m and 20 m spatial resolution bands with the latter upscaled to 10 m. Two sets of mosaics were created over the entire metropolitan French territory: a single leaf-on season mosaic (April-September 2020) for the pre-trained model, and three four-month mosaics (October 2019-September 2020) for the multitemporal AE. Cloud filtering was applied using the S-2 Cloud Probability dataset.

S-1 data were obtained from the S-1A and S-1B satellites operating at C-band, collected in Interferometric Wide swath mode with VV and VH polarizations. The data was calibrated, geometrically corrected, and resampled to 10 m resolution. Similar to S-2, we created two sets of mosaics with normalized backscattering coefficients.

GEDI is a full-waveform LiDAR sensor on the International Space Station (ISS), operational between 51.6$^{\circ}$ N and 51.6$^{\circ}$ S from 2019-2023. We used Level 1B, 2A, and 2B data from April 2019 to December 2021, extracting waveforms, geolocation, height metrics ($\mathcal{W}_{rh}$), canopy cover ($\mathcal{W}_{c}$), and Plant Area Index ($\mathcal{W}_{pai}$). After quality filtering following Fayad et al. \yrcite{Fayad_etal_2024} and seasonal selection, the dataset contained $\approx$ 19 million waveforms, covering less than 1\% of the total surface area of France. 

Overall, our pre-training dataset consisted of 836K $64 \times 64$ pixels S-1 \& 2 images with, on average, 26 GEDI waveforms per image. Additional details on the pre-training datasets can be found in \cref{appendix:experimental_settings}.

\paragraph{Evaluation Datasets.}

We evaluated our model on seven downstream tasks using labels from various data products at resolutions matching or lower than our model's output (10 m). 
\emph{PureForest ($PF$)} provides a benchmark dataset of ground truth patches for classifying mono-specific forests in France, featuring high-resolution imagery and annotations for over 135K $50 \times 50$ m (i.e., 5x5 pixels) patches across 13 tree species \cite{gaydon2024pureforest}. \emph{CLC+Backbone ($CLS_+$)} is a pan-European land cover inventory for 2021, utilizing S-2 time series (2020-2022) and a TempCNN classifier \citep{Pelletier2019} to produce a 10 m raster indicating the dominant land cover among 11 classes. \emph{PASTIS} is a crop mapping dataset by Garnot et al. \yrcite{garnot2021pastis}, covering 18 crop classes and 1 background class with 2433 densely annotated $128 \times 128$ pixels images at 10 m resolution. The \emph{vertical structure dataset} assesses model performance in mapping forest heights, canopy cover, plant area index, and waveform retrieval at 10 m resolution, relying on GEDI-derived products as presented earlier. When available, we used the train/val/test split used by the references (i.e., $PF$ and \emph{PASTIS}). For the $CLS_+$ dataset, we used the same split as the unsupervised dataset, which followed a 65/10/25 split. 

\subsubsection{Performance Evaluation}
We evaluated our model's retrieval capacities for both zero-shot dense prediction tasks and its performance in the fine-tuned setting. For both settings, we used the weighted F1 score (\textit{w}F1) for classification tasks, root mean squared error (RMSE) and Pearson's correlation coefficient ($r$) for regression tasks. To evaluate the similarity between acquired and retrieved/generated waveforms, we used Pearson's correlation coefficient, computed between the time-aligned retrieved/generated waveforms and the acquired waveforms.

For zero-shot classification, we constructed retrieval databases based on the downstream tasks. For vertical structure tasks, outputs from $\mathcal{O}^\mathcal{V}$ served as queries, and we created a single database with $L_2$ normalized waveform embeddings from $\mathcal{O}^\mathcal{W}$ as keys, paired with four target labels: $\mathcal{W}_{rh}$, $\mathcal{W}_c$, $\mathcal{W}_{pai}$, and the complete waveform $(\mathcal{W})$. For horizontal structure tasks, outputs from $\mathcal{O}^\mathcal{H}$ were used as queries, creating a separate database for each target as not all targets were available at all pixels simultaneously. Here, keys are $L_2$ normalized pixel embeddings from $\mathcal{O}^\mathcal{H}$, with targets being a land cover class (i.e., $CLC_+$), crop type (i.e., $PASTIS$), or tree species (i.e., $PF$). For tree species identification, the labels cover a $5 \times 5$ pixel area, so queries and keys are the averaged embeddings over this window. Next, given an input image and its $L_2$ normalized pixel embeddings from $\mathcal{O}^\mathcal{V}$ or $\mathcal{O}^\mathcal{H}$, we retrieve the $k$ nearest neighbors (KNN) for each pixel based on cosine similarity and assign the target class by distance-weighted voting. For the KNN retrieval, keys were obtained from the training split while the queries were obtained from the test split.  

For the low-shot fine-tuning, we froze the entire pre-trained network except for the last NA layer in each decoder, which were appended with an output head consisting of two sequential $1 \times 1$ convolutional layers. The first layer halves the input channels, and the second projects the reduced representation to the desired output size.

We evaluated zero-shot classification and low-shot fine-tuning in a low-data regime. We define a dataset with a low number of labels based on the type of labels usually available for this dataset. For $CLC_+$ and $PASTIS$, labeled data are available as densely annotated images. For $PF$, $\mathcal{W}$, $\mathcal{W}_{rh}$, $\mathcal{W}_c$ and $\mathcal{W}_{pai}$, labeled data are available as single annotated pixels.

\subsubsection{Competing Models.} 
We compared DUNIA in the fine-tuned setting to five current state-of-the-art Earth Observation FMs: SatMAE \cite{cong2022satmae}, DOFA \cite{Xiong_etal_2024}, DeCUR \cite{wang2024decur}, CROMA \cite{fuller2023croma} and AnySat \cite{astruc2024anysat}. SatMAE takes as input S-2 imagery, DOFA, DeCUR, and CROMA take as input S-1 \& 2 imagery, while Anysat is pre-trained using S-1 \& 2 times series as well as very high-resolution imagery. For a fair comparison, we pre-trained all competing models (using pre-trained weights when available) for 250K steps using our datasets and the training details from the respective papers. For AnySat, we used multi-temporal S-1 \& 2 mosaics with three time steps and included SPOT images at 1.5 m resolution as an additional input modality during pre-training but fine-tuned using only S-1 \& 2. All competing models were evaluated on all downstream tasks except waveform generation, as they do not support this task.

\begin{table*}[!ht]
\caption{Top-1 retrieval-based zero-shot classification performance of DUNIA for different retrieval settings. SOTA results, when available, represent the best-performing supervised models for any given dataset. \textsuperscript{\textdagger} are retrained models for a particular dataset. \colorbox{cgreen!50}{\textcolor{cgreen!50}{---}} and \colorbox{cblue!50}{\textcolor{cblue!50}{---}} are DUNIA query embeddings from $\mathcal{O^V}$ and $\mathcal{O^H}$ respectively. S represents the number of samples in the retrieval database, with \textit{im} meaning a $64\times64$ pixels fully annotated image, and \textit{l} meaning a single annotated pixel. KNN represents the $k$ nearest neighbors used in the distance-weighted voting. $\mathcal{W}^{**}$ represents performance results for vertical structures higher than 5 m. Best scores are in \textbf{bold}.}
\label{table:retrieval_performance}
\begin{center}
\begin{small}
\begin{sc}
\begin{tabular}{lcccccc}
\toprule
\multirow[t]{2}{*}{Dataset} & \multirow[t]{2}{*}{Metric} & \multirow[t]{2}{*}{SOTA} & \multirow[t]{2}{*}{S} & \multicolumn{3}{c}{DUNIA}                                  \\
\cmidrule(lr){1-1} \cmidrule(lr){2-2} \cmidrule(lr){3-3} \cmidrule(lr){4-4} \cmidrule(lr){5-7}                       
                         &            &     &       & KNN = 50 & KNN = 5 & KNN =1 \\
\midrule                         
\cellcolor{cgreen!50}& &\multicolumn{1}{l}{\citet{schwartz2023forms}} 5.2 (.77) & & & &\\ 
\cellcolor{cgreen!50}& &\multicolumn{1}{l}{\citet{liu2023overlooked}} 5.2 (.76) & & & &\\
\cellcolor{cgreen!50}& &\multicolumn{1}{l}{\citet{Tolan_etal_2024}} 8.5 (.59) & & & &\\
\multirow{-4}{*}{\cellcolor{cgreen!50}{$\mathcal{W}_{rh}$} }                 & \multirow{-4}{*}{\textnormal{RMSE (r)}}            &\multicolumn{1}{l}{\citet{Lang_etal_2023}} 5.6 (.76)         & \multirow{-4}{*}{50K \textit{l} $^*$}    & \multirow{-4}{*}{\textbf{2.0} (.93)}        & \multirow{-4}{*}{2.1 (.93)}        & \multirow{-4}{*}{2.1 (.91)}    \\

\cellcolor{cgreen!50}$\mathcal{W}_c$                    &\textnormal{RMSE (r)}            &  \multicolumn{1}{l}{\citet{schwartz2023forms}\textsuperscript{\textdagger}} 22.1 (.54)        &50K \textit{l}      &\textbf{11.7} (.89)       &12.1 (.84)        &12.3 (.86)      \\ 
\cellcolor{cgreen!50}$\mathcal{W}_{pai}$                &\textnormal{RMSE (r)}            &  \multicolumn{1}{l}{\citet{schwartz2023forms}\textsuperscript{\textdagger}}  1.5 (.35)       &50K \textit{l}      & \textbf{0.71} (.75)      &0.74 (.73)        &0.74 (.72)     \\ 
\cellcolor{cblue!50}$CLC_+$                  &\textnormal{\textit{w}F1}            &  ---        &500 \textit{im}      &\textbf{80.1}        &74.3 &69.5    \\ 
\cellcolor{cblue!50}$PASTIS$                 &OA             & \multicolumn{1}{l}{\citet{garnot2021pastis}} \textbf{84.2}          &500 \textit{im}      &56.2        &54.1  &49.5    \\ 
\cellcolor{cblue!50}$PF$                     &\textnormal{\textit{w}F1}            & \multicolumn{1}{l}{\citet{gaydon2024pureforest}} 74.6          &50K \textit {l}      &73.8        &\textbf{76.0 }       &75.8    \\ 
\cellcolor{cgreen!50}$\mathcal{W}^{**}$            &\textnormal{r}           & ---          &50K \textit{l}      & ---       &---        &\textbf{.70} \\
\bottomrule
\end{tabular}%
\end{sc}
\end{small}
\end{center}
\footnotesize{$^*$  This number of samples is considered as a low data regime due to: 1) a total coverage area of $\approx31Km^2$ 2) only $\approx0.25\%$ of data required compared to some supervised models.}
\end{table*} 

\subsection{Results}
Our evaluation shows that embeddings from our proposed framework, combined with simple zero-shot classifiers, surpass specialized supervised models in several tasks. In the fine-tuned setting, our model demonstrates strong low-shot performance, rivaling or exceeding state-of-the-art methods.

\subsubsection{Zero-Shot Classification Performance}

Zero-shot results in \cref{table:retrieval_performance} demonstrate that for vertical structure products ($\mathcal{W}_{rh}$, $\mathcal{W}_{c}$, $\mathcal{W}_{pai}$), DUNIA with KNN=50 outperforms specialist models. It achieves RMSE (r) improvements of 3.2 m (.16), 10.4\% (.35), and 0.79 (.4) for, respectively, canopy height, cover, and plant area index, and, matching the mapping quality of supervised methods (\cref{fig:result_maps_canopy_cover,fig:result_maps_canopy_height}). For tree species identification ($PF$), DUNIA slightly underperforms by 0.8\% at KNN=50 but exceeds the baseline by 1.4\% when decreasing the number of neighbors (KNN=5) used for the distance-weighted voting. For land cover mapping ($CLC_+$), DUNIA achieves strong performance with a \textit{w}F1 score of 80.1\%. However, for crop type mapping ($PASTIS$), DUNIA significantly underperforms compared to the state-of-the-art, which is likely due to the variability of the phenological cycles of crops that cannot be well captured by a single median composite.

For lower KNN values, the performance of vertical structure-related products and tree species identification remains stable. In contrast, land cover and crop type mapping experience significant performance drops, indicating that the model needs more samples for accurate classification. For retrieval databases with fewer samples (\cref{table:retrieval_performance_label_quantity}), vertical structure-related products ($\mathcal{W}_{rh}$, $\mathcal{W}_c$, $\mathcal{W}_{pai}$) maintain their performance and map qualities (\cref{fig:high_vs_low_samples}), while other products ($CLC_+$, $PASTIS$, $PF$) show a substantial degradation in performance.

Regarding waveform ($\mathcal{W}$) retrieval performance, we observe that our model is able to retrieve relevant waveforms given a pixel embedding as a query, with a correlation coefficient of .70 for S=50k \textit{l}, and decreases by a small factor with smaller retrieval database sizes (\cref{table:retrieval_performance_label_quantity}). This is to be expected given the small subset of waveforms in the retrieval database that might not reflect all possible variations for a particular height class. Retrieved vs. reference waveforms can be found in \cref{fig:result_waveforms,fig:result_waveforms_set_2,fig:result_waveforms_set_3,fig:result_waveforms_set_4}.  

\subsubsection{Fine-tuning Performance}

\begin{table*}[!ht]
\caption{Fine-tuning performance of DUNIA and the five competing models. \colorbox{cgreen!50}{\textcolor{cgreen!50}{---}} and \colorbox{cblue!50}{\textcolor{cblue!50}{---}} are DUNIA's embeddings from $\mathcal{O^V}$ and $\mathcal{O^H}$ respectively. S represents the number of samples used for the fine-tuning, with \textit{im} meaning a $64\times64$ pixels fully annotated image, and \textit{l} meaning a single annotated pixel. $\mathcal{W}^{**}$ represents performance results for vertical structures higher than 5 m. Best scores are in \textbf{bold}.}
\vspace{-1em}
\label{table:finetune_performance}
\vskip-0.25cm
\begin{center}
\begin{small}
\begin{sc}
\resizebox{\textwidth}{!}{%
\begin{tabular}{lccccccccc}
\toprule
Dataset & Metric & Samples (S) & DUNIA & AnySat & Croma & DOFA & DeCUR & SatMAE \\
\midrule
\cellcolor{cgreen!50}$\mathcal{W}_{rh}$                 &\textnormal{RMSE (r)}    &5K \textit{im} &\textbf{1.3} (.95) &2.8 (.89) &3.5 (.78)  &11.0 (.51)  &11.0 (.55) &10.5 (.52) \\
\cellcolor{cgreen!50}$\mathcal{W}_c$                    &\textnormal{RMSE (r)}    &5K \textit{im}&\textbf{9.8} (.85) &12.1 (.79) &14.2 (.73) &29.2 (.50) &28.5 (.48) &30.2 (.48) \\
\cellcolor{cgreen!50}$\mathcal{W}_{pai}$                &\textnormal{RMSE (r)}    &5K \textit{im}&\textbf{0.62} (.71) &0.95 (.67) & 1.2 (.61) &1.5 (.38) &1.6 (.38)  &1.6 (.39) \\
\cellcolor{cblue!50}$CLC_+$                  &\textnormal{\textit{w}F1}         &5K \textit{im} &\textbf{90.3} &90.1 &86.4  &72.0  &75.1 &75.0 \\
\cellcolor{cblue!50}$PASTIS$                 &\textnormal{\textit{w}F1}          &1.5K \textit{im} &77.0 &\textbf{81.1} &73.3  &54.5   &57.3  & 55.2 \\
\cellcolor{cblue!50}$PF$                     &\textnormal{\textit{w}F1}         &50K \textit{l}&82.2 &\textbf{82.3} &80.5  &79.8  &78.9 &78.8 \\
\cellcolor{cgreen!50}$\mathcal{W}^{**}$ & \textnormal{r} & $\approx$19M \textit{l} & .78 &--- &--- &--- &--- &--- \\
\bottomrule
\end{tabular}%
}
\end{sc}
\end{small}
\end{center}
\vskip-0.3cm
\end{table*} 

Fine-tuning results in \cref{table:finetune_performance} demonstrate that DUNIA outperforms the other models in vertical structure-related products, with all models except AnySat trailing significantly. In comparison to DUNIA's zero-shot results, all the models underperformed in the finetuning setting for $\mathcal{W}_c$ and $\mathcal{W}_{pai}$ due to the long-tailed distribution of these two products that are not well modeled with the simple $L_1$ loss that we used. Qualitatively, as shown in \cref{fig:result_maps_canopy_cover,fig:result_maps_canopy_height}, DUNIA and AnySat show a similar level of detail; however, DUNIA is capable of producing higher values, which we attribute to the alignment with the vertical structure in the pre-training stage. For CROMA, \cref{fig:result_maps_canopy_cover,fig:result_maps_canopy_height} shows that this model produces much smoother maps than the two other models.

For $CLC_+$ and $PF$, DUNIA's performance in the fine-tuned setting increased in comparison to the zero-shot setting with results on par with the much more data-heavy AnySat model (\cref{table:finetune_performance}). However, DUNIA underperformed by a significant margin ($\downarrow4.1\%$) in comparison to AnySat for the $PASTIS$ dataset. Qualitatively (\cref{fig:result_maps_clc}) both DUNIA and AnySat show similar level of detail for the $CLC_+$ dataset, in contrast to CROMA which also showed very smooth maps with fewer details.

In the novel task of waveform generation, we observed a correlation ($r$) increase of .08 between reference and generated waveforms compared to retrieval ($r$=.70 with S=50K \textit{l}, KNN=1). This correlation dropped to .75 when the diffusion model was trained on just 20\% of the available waveforms (\cref{table:finetune_performance_label_quantity}). Retrieved vs. generated waveforms can be found in \cref{fig:result_waveforms,fig:result_waveforms_set_2,fig:result_waveforms_set_3,fig:result_waveforms_set_4}.

To further analyze the contributions of different architectural choices, including the use of separate decoders and the impact of our alignment strategies, we provide additional ablation studies in \cref{appendix:ablations}.

\section{Discussion}
We introduced DUNIA, a novel framework for Earth Observation applications that learns dense representations through mono-modal and cross-modal alignment. This strategy enables our model to perform mono-modal and cross-modal retrieval tasks on several datasets with high precision. It also allows novel capabilities like retrieving or generating highly complex outputs from pixel inputs. To the best of our knowledge, this is the only model capable of generating GEDI waveforms at this scale. 

Our main contributions lie in several design choices at the decoding stage. The proposed pre-trained model encoder is flexible and can be adapted to components from other existing approaches; for instance, modules that support multi-temporal input, or handle images with varying resolutions. Nonetheless, in this work we focused on input data - S-1 \& 2, alongside GEDI - that are freely accessible globally to promote easier adoption.   

One of the main limitations of this work is also coincidentally the reliance on static inputs from S-1 \& 2 data, which has limitations for datasets requiring multi-temporal images, such as crop type mapping. However, our design choices prioritize mono-temporal data to reduce storage requirements and enable applications in areas where time series data are not readily available. Another limitation, common to all EO models, is that the current pre-trained model is tailored to the area and year on which it was trained on; we expect that it will require pre-training when used elsewhere. Nonetheless, the model was designed to be pre-trained with limited computational resources. 

\section*{Acknowledgements}
This work is supported via the AI4Forest project, which is funded by the French National Research Agency (ANR; grant number ANR-22-FAI1-0002) and the German Federal Ministry of Education and Research (BMBF; grant number 01IS23025A).

\section*{Impact Statement}
Forests are the third-most important terrestrial carbon sink for climate change mitigation by absorbing about one-third of anthropogenic $\text{CO}_2$ emissions~\cite{friedlingsteinGlobalCarbonBudget2022}. However, these ecosystems are under increasing threat from deforestation, degradation, and climate-induced disturbances such as wildfires, insect outbreaks, and droughts \cite{anderegg2022climate}. Global initiatives, including the Glasgow Declaration and the Sustainable Development Goals\footnote{\url{https://sdgs.un.org/2030agenda}} emphasize the need for accurate forest monitoring to support conservation, restoration, and carbon sequestration projects. Unfortunately, despite these high praises, many other major initiatives have struggled due to a lack of reliable, high-resolution data on forest carbon stocks and how they change over time.

Current forest monitoring systems heavily rely on ground-based inventories, which provide robust statistical estimates of biomass and carbon at national or regional scales but lack the spatial granularity needed for tracking localized carbon gains and losses. Additionally, the largest forested regions—particularly in boreal and tropical areas—remain vastly under-sampled. While satellite data offers global coverage, traditional approaches are often constrained by their reliance on extensive labeled datasets, limiting their ability to map forest properties comprehensively.

DUNIA introduces a novel self-supervised learning approach that overcomes these limitations by generating pixel-level embeddings from freely accessible satellite data, including optical and radar imagery. By aligning these embeddings with sparse but highly informative LiDAR waveforms, our approach enables the estimation of multiple forest attributes—including canopy height, fractional cover, land cover, tree species, and vertical structure—without requiring task-specific labels. This method allows for zero-shot and low-shot predictions of key forest variables, paving the way for scalable monitoring of global ecosystems.

By providing a unified, multimodal representation of forests, DUNIA democratizes access to advanced Earth Observation tools, enabling researchers, policymakers, and conservationists to make informed decisions with minimal reliance on expensive field data collection. This work lays the foundation for high-resolution, large-scale ecosystem monitoring, offering a transformative approach to quantifying forest structure, biomass, and biodiversity in support of climate change mitigation and sustainable land management.

\bibliography{references}
\bibliographystyle{icml2025}

\newpage
\appendix
\onecolumn
\counterwithin{figure}{section}
\counterwithin{table}{section}

\renewcommand{\thetable}{\Alph{section}\arabic{table}}
\renewcommand{\thefigure}{\Alph{section}\arabic{figure}}

\section{Implementation Details}\label{appendix:implementation_details}

In the main text, we provided a high-level overview of the architecture and training setup. Here, we include additional implementation details for key modules and components of our framework. These details were omitted from the main body for brevity and are provided here for completeness and reproducibility.

\begin{figure*}[ht!]
\begin{center}
\includegraphics[width=\textwidth]{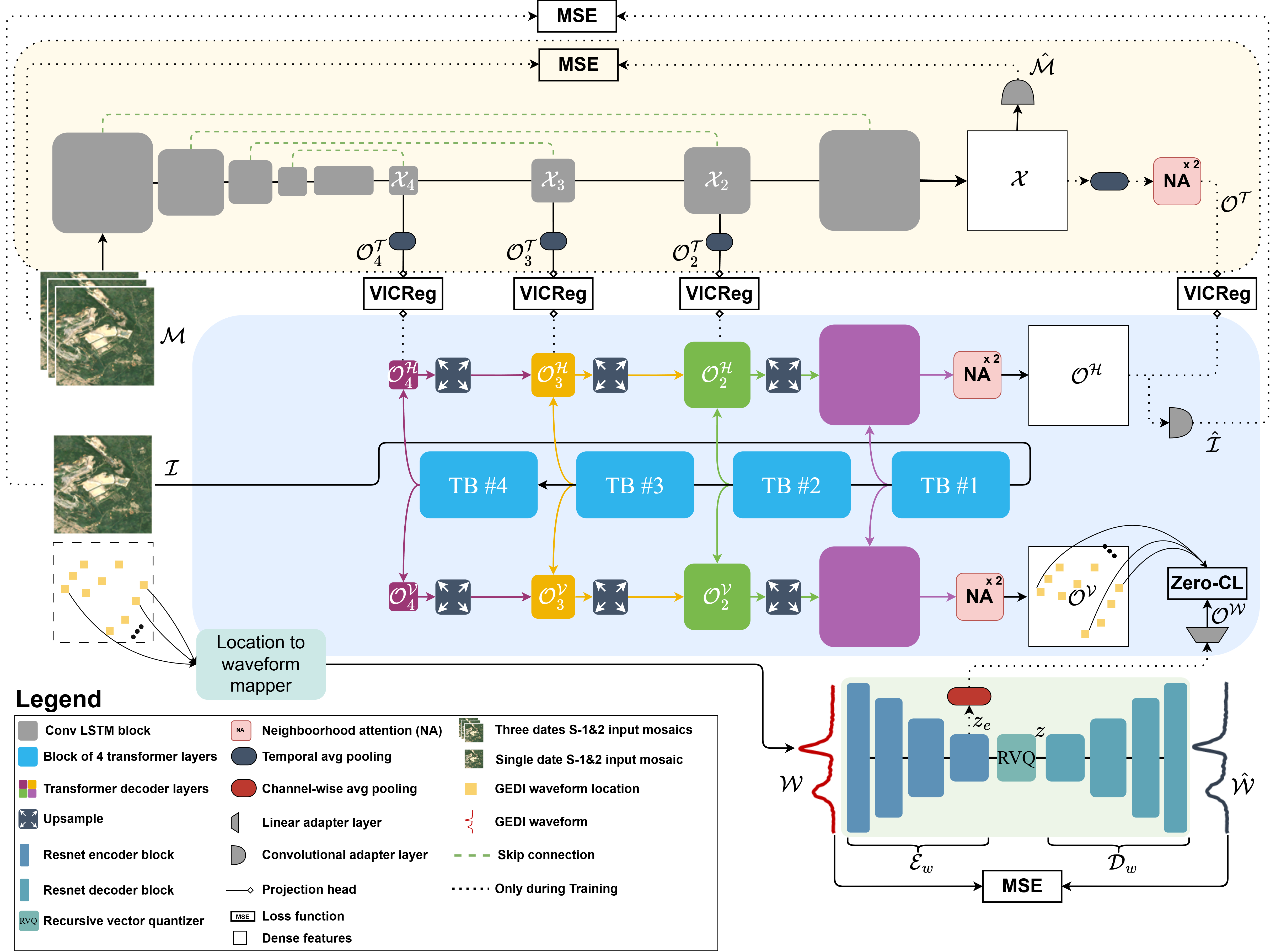}
\end{center}
\caption{An overview of our framework showcasing the main blocks and connections. The pre-trained model, which take as input a single date S-1 \& 2 image mosaic, is shaded in blue. The multi-temporal image AE, which take as input a multi-date S-1\& 2 image mosaic, is shaded in yellow. The vertical structure AE, which take as input the GEDI waveforms, is shaded in green.
}
\label{fig:dunia_full_overview}
\vskip-0.14cm
\end{figure*}

\subsection{Pre-Trained Model}

\paragraph{Patch Embedding.} To generate patch embeddings, the 2D composite image $\mathcal{I} \in \mathbb{R}^{H \times W \times C}$ is processed using a convolutional layer with kernel size $(P, P)$ and stride $P$. This operation partitions $\mathcal{I}$ into $N = HW/P^2$ non-overlapping patches, which are reshaped into a sequence of embeddings in $\mathbb{R}^{N \times D}$. Here, $C$ represents the number of input channels, $(H, W)$ denotes the resolution of the input image, $P$ is the patch size, and $D$ is the patch embedding dimension.

Unlike tokenization strategies that rely on fixed-size input grids, the use of convolutions allows the model to adapt to varying input resolutions. This is particularly useful in EO applications, where spatial resolutions can vary across datasets. While transformer-based architectures address sequence length variability through techniques like rotary embeddings \cite{su2021roformer}, ALiBi \cite{press2022train}, and relative position encodings \cite{shaw2018self}, convolutional patch embedding provides a structured inductive bias that benefits spatial feature extraction while maintaining translational invariance.

\paragraph{Decoder Architecture.} The decoding process begins by reshaping the encoded token sequence from $\mathbb{R}^{N \times D}$ into an image-like feature representation at different resolutions. At each decoder stage $d \in {1,2,3,4}$, tokens are rescaled to $\frac{H}{2^{d-1}} \times \frac{W}{2^{d-1}} \times D_{p_d}$, where $D_{p_d} = 2^{d-1}D_p$ represents the channel dimension at each stage. This transformation involves an initial $1 \times 1$ convolution to adjust channel dimensions, followed by strided $3 \times 3$ convolutions for downsampling when $2^{d-1} \geq P$, or transpose convolutions when $2^{d-1} < P$.

To reconstruct high-resolution embeddings, feature maps from deeper layers ($d=4$) are progressively upsampled using bilinear interpolation and refined through two $3 \times 3$ Conv-BatchNorm-ReLU operations before being concatenated with corresponding features from earlier transformer layers. A final $1 \times 1$ convolution ensures that the final output dimension matches $D_p$ when required. This architecture is instantiated as two separate decoders, outputting $\mathcal{O}^\mathcal{V}$ and $\mathcal{O}^\mathcal{H}$.

\paragraph{Projection Head.} For the pixel-pixel alignment objective, we apply a lightweight two-layer $1 \times 1$ convolutional projection head to each output $\mathcal{O}^\mathcal{H}_d$. The first layer expands the feature dimension by a factor of two, followed by a second layer that projects it back to $D_p$. This design enhances feature expressiveness during training but, as observed in prior work \cite{xue2024projectionhead}, projection heads tend to overfit to the pretraining task, reducing their transferability to downstream applications. Consequently, we discard the projection head after training. In contrast, the pixel-waveform alignment output $\mathcal{O}^\mathcal{V}$ is directly used without an additional projection head, as its representation remains well-aligned with downstream tasks such as waveform retrieval and generation.

\subsection{Waveform Model}
\paragraph{Waveform Encoder.} The waveform encoder consists of an input stem and three stages designed to progressively reduce the 1D waveform of length $L$ (i.e., $\mathcal{W} \in \mathbb{R}^{L \times 1}$) while increasing its feature representation capacity. First, the input stem processes the waveform using a $3 \times 1$ convolutional layer with a stride of 2. This initial operation halves the length of the waveform to \( L/2 \) and increases the channel dimension to 2. Following the input stem, the encoder comprises three stages, each containing three ResNet blocks. Each ResNet block is designed with two $3 \times 1$ convolutional layers, batch normalization, ReLU activations, and a skip connection. At the beginning of each stage, the first ResNet block incorporates a strided convolution in its initial layer to halve the waveform’s length and double the channel dimension. The subsequent blocks within the same stage operate with a stride of 1. By the end of the three stages, the waveform encoder reduces the input waveform’s length to \( L/16 \) while progressively increasing the channel dimension to 16. Thus, the waveform encoder $\mathcal{E}_w$ transforms a waveform of shape $\mathbb{R}^{L \times 1}$ into a latent representation $z \in \mathbb{R}^{L/16 \times 16}$.

The alignment of the waveforms and the pixels is performed on this latent representation $z_e = \mathcal{E}_w(\mathcal{W})$. However, to adapt them to the outputs of $\mathcal{O}^\mathcal{V}$, we first apply an average pooling along the channel dimension, followed by a projection to the $\mathcal{O}^\mathcal{V}$ projection dimension (i.e., $D_p$) using a linear layer. In essence, this process transforms $z_e$ into $\mathcal{O}^\mathcal{W} \in \mathbb{R}^{D_p}$.

\paragraph{Residual Vector Quantizer (RVQ).} The RVQ layer takes a segment from the latent waveform representation $z^i_e \in \mathbb{R}^{1/16}$, where $z^i_e$ is the $i^{th}$ row of the waveform latent representation, and iteratively quantizes it. It operates by first mapping each $z^i_e$ to the closest codebook vector in the first quantizer, then computing the residual (i.e., the difference between the input vector and the quantized approximation). This residual is passed to the next quantizer in the sequence, and the process is repeated for all $Q$ quantizers. At the end of the process, the latent waveform representation $z_e$ is converted into a discrete series of quantized residual vectors, which are summed to produce the final quantized waveform representation $z$. 

\paragraph{Waveform Decoder.} The waveform decoder ($\mathcal{D}_w$) mirrors the structure of the encoder but operates in reverse order to reconstruct the waveform from its quantized latent representation $z$. Similar to the encoder, the decoder consists of three stages and an output stem. However, instead of using strided convolutions for downsampling, the decoder replaces them with transpose convolutions to progressively upsample the waveform. Each stage in the decoder increases the length of the waveform while halving the channel dimension, reversing the transformations applied by the encoder. By the end of the decoding process, the latent representation is transformed back into a waveform $\hat{\mathcal{W}}$ with length \( L \).

\subsection{Multitemporal Image Processing Model}

\paragraph{Multitemporal Image AE.} The AE has four processing and downsampling stages. In each stage we replace the conventional double Conv2D-BatchNorm-ReLU layers in UNets with a two-layer ConvLSTM followed by a BatchNorm layer and ReLU activation layer. This design choice is to explicitly model temporal correlations across time steps as mentioned earlier. Downsampling is performed using a strided Conv3D layer applied on the spatial dimension, while upsampling is performed using a trilinear upsampling operation also on the spatial dimension, followed by a Conv3D layer. As such, the model takes a multi-temporal input image $\mathcal{M} \in \mathbb{R}^{T \times H \times W \times C}$, where $T$ is the temporal dimension, and produces a feature map $\mathcal{X} \in \mathbb{R}^{T \times H \times W \times D_p}$. Finally, as this is an autoencoder, the original input shape is recovered using a Conv3D layer producing $\mathcal{\hat{M}} \in \mathbb{R}^{T \times H \times W \times C}$. All Conv3D layers in the AE have a kernel of $3 \times 3$ on the spatial dimension.
\section{Objective Functions.} \label{appendix:objective_functions}

\paragraph{Zero-CL Loss.}
Let \( Z^A \in \mathbb{R}^{G\times D_p} \) represent a $L_2$ normalized pixel embeddings from $\mathcal{O}^\mathcal{V}$, where \( G \) is the number of waveforms in a minibatch and \(D_p \) is the feature dimension, and \( Z^B \in \mathbb{R}^{G \times D_p} \) denote their corresponding $L_2$ normalized waveform embeddings from $\mathcal{O}^\mathcal{W}$. The instance-wise contrastive loss is defined as:

\begin{equation}
    \label{zero_icl}
    \mathcal{L}_{\text{Ins}} = \sum_{i=1}^{G} \left( 1 - \sum_{d=1}^{D_p} H^{A,\text{Ins}}_{i, d} \cdot H^{B,\text{Ins}}_{i, d} \right)^2
\end{equation}
\( H_{i,d} \) represents the $d^{th}$ feature value of the $i^{th}$ instance. $H^{\text{Ins}}$ is a zero-phase component analysis (ZCA) whitened embedding matrix defined as:
\begin{equation}
    \label{zca_whitening_ins}
    H^{\text{Ins}} = W^{\text{Ins}} Z, \quad W^{\text{Ins}} = E^{\text{Ins}} \Lambda_S^{-1/2} E^\top
\end{equation}
where \( E^{\text{Ins}} \in \mathbb{R}^{G,G}\) and \( \Lambda_S \) are the eigenmatrix and diagonal of the eigenvalue matrix of the \emph{affinity matrix} $S=\frac{1}{D_p}ZZ^\top$, respectively.
Similar to the instance-wise contrastive objective, the feature-wise contrastive loss is formulated as:

\begin{equation}
    \label{zero_fcl}
       \mathcal{L}_{\text{Fea}} = \sum_{d}^{D_p} \left( 1 - \sum_{i}^{G} H^{A,\text{Fea}}_{i, d} \cdot H^{B,\text{Fea}}_{i, d} \right)^2
\end{equation}
Where $H^{\text{Fea}}$ is defined as:
\begin{equation}
    \label{zca_whitening_fea}
    H^{\text{Fea}} = W^{\text{Fea}} Z^\top, \quad W^{\text{Fea}} = E \Lambda_C^{-1/2} E^\top
\end{equation}
where \( E \in \mathbb{R}^{D_p,D_p}\) and \( \Lambda_C \) are the eigenmatrix and diagonal of the eigenvalue matrix of the \emph{covariance matrix} $C=\frac{1}{G}Z^\top Z$, respectively.

The overall loss $\mathcal{L}_{Zero-CL}$ is $\mathcal{L}_{\text{Fea}} + \mathcal{L}_{\text{Ins}}$.

\paragraph{VICReg Loss.}
Let \( Z^\mathcal{H} \in \mathbb{R}^{M\times D_{p}} \) represent an $L_2$ normalized pixel embedding from \( \mathcal{O}^\mathcal{H}\) and \( Z^\mathcal{T} \in \mathbb{R}^{M \times D_{p}} \) its corresponding $L_2$ normalized pixel embeddings from \( \mathcal{O}^\mathcal{T}\), with \( M = B \times H \times W \) the number of pixels in the mini-batch of size $B$.
VICReg is expressed as a sum of three losses: a variance loss, an invariance loss, and a covariance loss. The variance loss is formulated as:
\begin{equation}
    \label{var_loss}
    \begin{split}
    & \mathcal{L}_{\text{var}} = \frac{1}{2} (v(Z^\mathcal{H})+v(Z^\mathcal{T})),\\
    & v(Z) = \frac{1}{D_p}\sum_d^{D_p}\max\left(0,1-\sqrt{\frac{1}{M}\sum_i^M(Z_{i,d}-\overline{Z_d})^2+\epsilon}\right)
    \end{split}
\end{equation}
where \( Z_{i,d} \) represents the $d^{th}$ feature value of the $i^{th}$ instance.

The invariance loss is formulated as:

\begin{equation}
    \label{inv_loss}
    \mathcal{L}_{\text{inv}}= \frac{1}{M} \sum_{i=1}^M \| Z_i^H - Z_i^T \|_2^2 
\end{equation}
The covariance loss is formulated as:
\begin{equation}
    \label{cov_loss}
    \begin{split}
    & \mathcal{L}_{\text{cov}} = c(Z^\mathcal{H}) + c(Z^\mathcal{T}), \\
    & c(Z) = \frac{1}{D_p}\sum_{k\neq l}[cov(Z)]^2_{k,l} \\
    & cov(Z) = \frac{1}{M-1}\sum_{i}^M(Z_i-\overline{Z})(Z_i-\overline{Z})^\top
     \end{split}
\end{equation}
The overall loss term $\mathcal{L}_{\text{VICReg}}$ is then:
\begin{equation}
    \label{loss_vicreg}
    \mathcal{L}_{\text{VICReg}} = \alpha_v \mathcal{L}_{\text{var}} + \beta_i\mathcal{L}_{\text{inv}} + \gamma_c \mathcal{L}_{\text{cov}}
\end{equation}
Following Bardes \yrcite{bardes2021vicreg} we set $\alpha_v$ and $\beta_i$ to 25.0 and $\gamma_c$ to 1.0.

\section{Waveform Generation}\label{appendix:diffusion_math}
Denoising diffusion models (DM) \cite{sohl2015dm} are generative models that learn a data distribution \( p_{\text{data}} \) by gradually denoising a noisy variable following a predefined noise schedule. Let \( \epsilon_\theta \) be a neural network. Its goal is to predict the original data \( x \) from a noisy version \( x_t \) generated by \( x_t = \alpha_t x_0 + \sigma_t \epsilon \), where \( \epsilon \sim \mathcal{N}(0, \mathbf{I}) \) is Gaussian noise, \( \alpha_t \) and \( \sigma_t \) are parameters of a noise scheduling function, and $t$ is a time step from $\{1,...,T\}$.

The corresponding objective of a DM can then be written as:

\begin{equation}
    L_{DM} = \mathbb{E}_{x_0 \sim p_{data}, \epsilon \sim \mathcal{N}(0,\mathbf{I}), t} \left[ \left\| y - \epsilon_\theta(x_t, t) \right\|_2^2 \right]
\end{equation}

where $p_{data}$ denotes the data distribution over the clean inputs $x_0$, and the target $y$ can be the input noise $\epsilon$, the original input $x$ variable, or the velocity $v = \alpha_t \epsilon - \sigma_t x$. DMs can also be parametrized by a condition $c \in \mathbb{R}^D$ in addition to the diffusion time $t$, and the corresponding objective becomes:
\begin{equation}
    L_{CDM} = \mathbb{E}_{x_0 \sim p_{data}, \epsilon \sim \mathcal{N}(0,1), t,c} \left[ \left\| y - \epsilon_\theta(x_t, t,c) \right\|_2^2 \right]
\end{equation}

Another class of diffusion models, known as Latent Diffusion Models (LDMs) \cite{rombach2022sd,shen2023naturalspeech2,ramesh2022hierarchical}, operate in a latent space rather than directly in the high-dimensional data space. By operating in a lower-dimensional latent space, LDMs significantly reduce the computational cost of training and inference compared to DMs, which operate in the original data space.  LDMs leverage the latent representation of data obtained by training an autoencoder, defined by an encoder $\mathcal{E}$ and a decoder $\mathcal{D}$. The autoencoder is trained such that $\hat{x} = \mathcal{D}(\mathcal{E}(x))$ is a reconstruction of the original data $x$. After training the autoencoder, a diffusion model is trained directly on the encoded data samples $z = \mathcal{E}(x)$. A new sample is then obtained by first sampling a representation $z$, and then $\mathcal{D}(z)$ yields $x$. 

Following the LDM formulation, we learn a network $\epsilon_\theta$ conditioned on $\mathcal{O}^V_{\phi,\lambda}$, where $\phi$ and $\lambda$ are the waveform coordinates, that directly predicts a denoised waveform latent $z_0 \in \mathbb{R}^{\frac{L}{16}\times 16}$, where $z_0 = \mathcal{E}(\mathcal{W})$, by minimizing:
\begin{equation}
    \label{eq:cond_ldm}
        L_{LDM} = \mathbb{E}_{z_0 \sim p(z_0), \epsilon \sim \mathcal{N}(0,1), t\sim U[1,T],\mathcal{O}^V_{\phi,\lambda}} \left[ \left\|\epsilon_\theta(z_t, t,\mathcal{O}^V_{\phi,\lambda})-z_0 \right\|_2^2 \right]
\end{equation}
In our formulation, the network $\epsilon_\theta(z_t,t,\mathcal{O}^V_{\phi,\lambda})$ is a 1D UNet model which takes the current noisy latent $z_t$, the time step $t$ as input, and the condition $\mathcal{O}^V_{\phi,\lambda}$ is integrated into the UNet via cross-attention at each layer. 
To sample $z_0$ from the diffusion model, we start from a Gaussian noise $z_T$ and gradually denoise it into samples $z_{T-1},...,z_0$ using the ordinary differential equation (ODE) solver proposed by Karras et al. \yrcite{karras2022elucidating}, as we found that it produces high-quality samples with fewer denoising steps. 
To make the sampled latent waveform $z_0$ better match its conditioning we use classifier-free guidance. Specifically, during training we randomly drop the conditioning for 50\% of the examples to make the model capable of conditional and unconditional denoising. In practice, we replace the conditioning signal with a random vector. Within the classifier-free guidance formulation, model predictions can be written as:

\begin{equation}
    \label{eq:classifier_free_guidance}
    \hat{\epsilon_\theta}(z_t,t,\mathcal{O}^V_{\phi,\lambda}) = \epsilon_\theta(z_t,t) + s\cdot(\epsilon_\theta(z_t,t,\mathcal{O^V}_{\phi,\lambda})-\epsilon_\theta(z_t,t)) 
\end{equation}
where $s$ is the guidance scale. Setting $s$ to 0 is equivalent to unconditional sampling, while setting it to $\geq 1$ has shown to produce more coherent but less diverse results. Here we set $s$ to 3.
Finally, to synthesize a waveform at any location $(\phi,\lambda)$ given a pixel embedding $\mathcal{O}^V_{\phi,\lambda}$ as a condition, we use eq. \ref{eq:classifier_free_guidance} to get $\hat{z_0}$, and leverage the frozen waveform decoder $\mathcal{D}_w$, where $\mathcal{D}_w(\hat{z_0})$ is a generated waveform.

\section{Visual Comparison: Pixel vs. Patch Embeddings}\label{appendix:patch_vs_pixel_embeddings}
\begin{figure*}[!h]
\begin{center}
\includegraphics[width=0.95\textwidth]{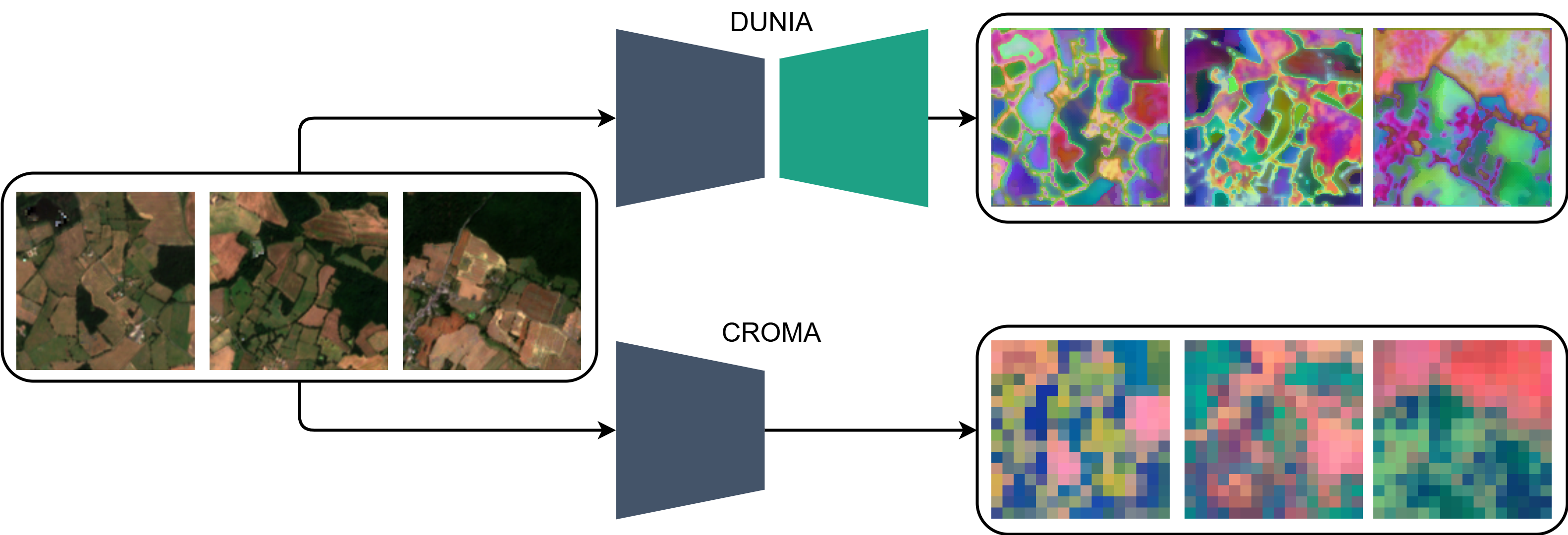}
\end{center}
\caption{Comparison between pixel-level embeddings from our proposed model (top row) and patch-based embeddings from one of the baseline models (bottom row), both generated from the same S-1 \& 2 inputs (left). Embeddings are visualized using t-SNE.
}
\label{fig:patch_vs_pixel_embeddings}
\end{figure*}

To highlight the difference in spatial granularity between our proposed pixel-level embedding model and a patch-based baseline (e.g., CROMA), we present a side-by-side qualitative visualization in Figure~\ref{fig:patch_vs_pixel_embeddings}. Both models are applied to the same S-1 \& 2 input, and their embeddings are projected to RGB using t-SNE. The pixel-based model captures finer spatial structures and preserves object boundaries more effectively than the patch-based baseline, yielding lower-resolution outputs.

\section{Model Configuration and Optimization}\label{appendix:model_config_and_optim}
\subsection{Model Configuration} The image model is pre-trained on $64 \times 64$ pixel sized images with 14 bands, a 16-layer transformer with 8 attention heads, a patch size of eight, 512 embedding dimensions, a feedforward layer with two linear layers with a hidden dimension of 2048, GeGLU activation, and 0.1 dropout and attention dropout rates. The two image decoders are identical and project the decoded image feature map to 64 embedding dimensions (i.e. $D_p = 64$). Finally, we set the window size $w$ of the NA layers to 19 pixels. The waveform AE is trained on 256 bin-long waveforms, with RVQ layers composed of eight quantizers, and each quantizer containing 512 codebook entries. As such, the waveform encoder encodes the waveforms into a $16 \times 16$ latent representation which is average pooled along the channel dimension and projected to 64 embeddings dimension for the pixel-waveform alignment. The multi-temporal image AE is trained on three $64 \times 64$ pixel sized input images, each with 14 bands, and produces a similarly shaped feature map on the temporal and spatial dimensions, and 64 embeddings dimension. This output is then average-pooled on the temporal dimension for the pixel-pixel alignment. Finally, the diffusion model is trained on the $16 \times 16$ quantized waveform latents, which are first projected to 128 channels before being passed to the UNet model. 

\subsection{Training, Inference, and Generation}
DUNIA was pre-trained on a single NVIDIA A6000 48GB GPU with a batch size of 60 for 250K steps using the Lion optimizer \cite{chen2024LION}, a learning rate of $5\mathrm{e}{-5}$, weight decay of $0.4$, 5K warmup steps, and a cosine annealing schedule. For finetuning, we used AdamW \cite{loshchilov2017adamw} with a $2\mathrm{e}{-4}$ learning rate, a cosine annealing schedule, a batch size of 20, and trained until plateau. The diffusion model was trained with a batch size of 4096 for 100K steps using AdamW, a $1\mathrm{e}{-4}$ learning rate, 5K warmup steps, and a cosine annealing schedule. During inference, diffusion steps were set to 30. Both DUNIA and the diffusion model were regularized with the Switch EMA (SEMA) technique \cite{li2024sema}, maintaining an EMA with a decay rate of 0.9, updating every 5 steps, and replacing online model parameters every 1K steps with a SEMA coefficient of 0.9.

\section{Experimental Settings}\label{appendix:experimental_settings}

\paragraph{Sentinel-2 Data.} We used Level-2A surface reflectance data (S2-L2A) from Google Earth Engine (GEE). The dataset includes bands at 10m (Blue, Green, Red, and NIR) and 20m (Red Edge 1-4, SWIR1, and SWIR2) spatial resolutions, all upscaled to 10m ground sampling distance (GSD) using cubic interpolation. We created two sets of mosaics: a single mosaic during the leaf-on season using the median of all available images from April to September 2020. This mosaic is used as input for the pre-trained model. The other set is used as input for the multitemporal AE and is comprised of three mosaics, each representing the median acquisitions of all images in a four-month span from October 2019 until September 2020. Cloudy pixels were filtered out using the S-2 Cloud Probability dataset provided by SentinelHub in GEE.

\paragraph{Sentinel-1 Data.} Sentinel-1 data were obtained from the Sentinel-1A and Sentinel-1B satellites, operating at the C-band (~6 cm wavelength). Images were collected in Interferometric Wide swath mode (IW) with VV and VH polarizations, derived from the high-resolution Level-1 ground range detected (GRD) product. The original 20 m × 22 m resolution was resampled to 10 m × 10 m GSD. Similar to S-2, we also created two sets of mosaics. The S-1 data were calibrated using the Sentinel SNAP toolbox, converting pixel values to backscattering coefficients ($\sigma^0$) in linear units and applying geometric correction using the 30m Shuttle Radar Topography Mission (SRTM) Digital Elevation Model (DEM). Finally, the backscattering coefficients $\sigma_\theta^0$ acquired at difference incidence angles $\theta$ were normalized to a common reference incidence angle set at $\sigma_{ref} = 40^\circ$ using the cosine correction equation (\citet{Baghdadi2001,TopouzelisSinghaKitsiou}):

\begin{equation}
    \sigma_ref^0 = \frac{\sigma_\theta^0 cos^2(\theta_{ref})}{cos^2(\theta)}
\end{equation}

\paragraph{GEDI Data.} GEDI measures the vertical structure of objects (e.g., vegetation) using three lasers emitting near-infrared light (1064 nm), one of the laser beams is split in two, and after applying optical dithering, this results in eight ground tracks spaced 600 m apart, with each shot having a 25 m footprint. For this study, we used GEDI Level 1B (L1B), Level 2A (L2A), and Level 2B (L2B) data from April 2019 to December 2021. From the L1B product we extracted the waveforms and their geolocation (i.e., latitude, longitude, and elevation), and the geolocation of the GEDI instrument for each shot. Finally, as the waveforms are stored as vectors of 1420 bins (1 bin = 0.15 m), we also extracted for each waveform the bin count, the signal start ($\mathcal{W}_{start}$) and signal end ($\mathcal{W}_{end}$). The latter two indicate the location of the canopy top and the end position of the waveform before the noise, respectively. From the L2A data product, we extracted the $RH_{98}$ (referred to $\mathcal{W}_{rh}$), which represents the estimated height of the tallest object within the waveform footprint, the canopy cover ($\mathcal{W}_{c}$). Finally, from the L2B data product, we extracted the Plant Area Index ($\mathcal{W}_{pai}$). 

As the waveforms are affected by the atmospheric conditions at acquisition time, we followed the filtering scheme of Fayad et al. \yrcite{Fayad_etal_2024} to remove low-quality waveforms. Finally, due to the different waveform shapes between the leaf-off and leaf-on seasons, especially for those acquired over deciduous forests, we only considered waveforms acquired during the leaf-on season. This resulted in a GEDI dataset of $\approx$19 million waveforms and their associated metrics. 

\paragraph{PureForest ($PF$).}The PureForest dataset was designed as a benchmark and provides ground truth patches for the classification of mono-specific forests in France \cite{gaydon2024pureforest}. It includes high-resolution imagery and corresponding annotations for more than 13K $50 \times 50$ m forest patches for 13 tree species.

\paragraph{CLC+Backbone ($CLS_+$).}The $CLS_+$ A pan-European wall-to-wall land cover inventory for the 2021 reference year. The product is based on Sentinel 2 (i.e., optical) time series from 2020 to 2022 and a temporal convolutional neural network (TempCNN) \cite{Pelletier2019} serving as a classifier. The product is available as a 10 m raster and shows for each pixel the dominant land cover among the 11 basic land cover classes.

\paragraph{PASTIS dataset.}A crop mapping dataset by Garnot \yrcite{garnot2021pastis} for 18 crop classes and 1 background class from the French Land Parcel Information System. The dataset contains 2433 $128 \times 128$ pixels densely annotated patches at 10 m resolution.

\paragraph{Vertical Structure dataset.}For vertical structure evaluation, we evaluated our model on its ability to map at 10 m resolution: (1) forest heights ($\mathcal{W}_{rh}$), forest fractional canopy cover ($\mathcal{W}_c$), plant area index ($\mathcal{W}_{pai}$), and complete waveform ($\mathcal{W}$) retrieval or generation. For these products, we rely on the products derived from the GEDI dataset presented earlier.

\section{Impact of Label Quantity on Performance}\label{appendix:label_quantity}

\begin{table}[h!]
\vspace{-1em}
\caption{Top-1 retrieval-based zero-shot classification performance of DUNIA for different database sizes. \colorbox{cgreen!50}{\textcolor{cgreen!50}{---}} and \colorbox{cblue!50}{\textcolor{cblue!50}{---}} are DUNIA query embeddings from $\mathcal{O^V}$ and $\mathcal{O^H}$ respectively. S represents the number of samples in the retrieval database, with \textit{im} meaning a $64\times64$ pixels fully annotated image, and \textit{l} meaning a single annotated pixel. $\text{KNN}_b$ represents the best KNN value for a given dataset. $\mathcal{W}^{**}$ represents performance results for vertical structures higher than 5 m.}
\label{table:retrieval_performance_label_quantity}
\begin{center}
\begin{small}
\begin{sc}
\begin{tabular}{lcccccc}
\toprule
Dataset & Metric & S & $\text{KNN}_b$ & 100\% S & 10\% S & 5\% S \\
\midrule                    
\cellcolor{cgreen!50}{$\mathcal{W}_{rh}$}                 &\textnormal{RMSE (r)}         &50K \textit{l} $^*$ &50    & 2.0 (.93)        &2.1 (.92)        & 2.1 (.92)    \\

\cellcolor{cgreen!50}$\mathcal{W}_c$                    &\textnormal{RMSE (r)}           &50K \textit{l} & 50     &11.7 (.89)       &12.4 (.86)        &12.0 (.84)      \\ 
\cellcolor{cgreen!50}$\mathcal{W}_{pai}$                &\textnormal{RMSE (r)}           &50K \textit{l} &50      & 0.71 (.75)      &0.72 (.75)        &0.72 (.74)     \\ 
\cellcolor{cblue!50}$CLC_+$                  &\textnormal{\textit{w}F1}                              &500 \textit{im} & 50     &80.1        &75.2 &70.2  \\ 
\cellcolor{cblue!50}$PASTIS$                 &OA             &500 \textit{im} & 50      &56.2        &52.4  &48.3    \\ 
\cellcolor{cblue!50}$PF$                     &\textnormal{\textit{w}F1}                      &50K \textit {l} & 5      &76.0        &73.5        &70.9  \\ 
\cellcolor{cgreen!50}$\mathcal{W}^{**}$            &\textnormal{r}           &50K \textit{l}  & 1    & .70       &.67        &.66 \\
\bottomrule
\end{tabular}%
\end{sc}
\end{small}
\end{center}
\end{table}

\vspace{-2em}
\begin{table}[H]
\caption{Fine-tuning performance of DUNIA and the five competing models. \colorbox{cgreen!50}{\textcolor{cgreen!50}{---}} and \colorbox{cblue!50}{\textcolor{cblue!50}{---}} are DUNIA's embeddings from $\mathcal{O^V}$ and $\mathcal{O^H}$ respectively. S represents the number of samples used for the fine-tuning, with \textit{im} meaning a $64\times64$ pixels fully annotated image, and \textit{l} meaning a single annotated pixel. $\mathcal{W}^{**}$ represents performance results for vertical structures higher than 5 m. Best scores are in \textbf{bold}.}
\label{table:finetune_performance_label_quantity}
\begin{center}
\begin{small}
\begin{sc}
\begin{tabular}{lccccccccc}
\toprule
Dataset & Metric & Samples (S) & DUNIA & AnySat & Croma & DOFA & DeCUR & SatMAE \\
\midrule                    
\cellcolor{cgreen!50}$\mathcal{W}_{rh}$                 &\textnormal{RMSE (r)}    &1K \textit{im} &\textbf{1.4} (.93) &2.8 (.89) &3.6 (.76)  &11.2 (.50)  &11.1 (.52) &10.5 (.52) \\
\cellcolor{cgreen!50}$\mathcal{W}_c$                    &\textnormal{RMSE (r)}    &1K \textit{im}&\textbf{10.0} (.83) &12.4 (.79) &14.5 (.72) &30.1 (.48) & 29.4 (.47) &30.7 (.47) \\
\cellcolor{cgreen!50}$\mathcal{W}_{pai}$                &\textnormal{RMSE (r)}    &1K \textit{im}&\textbf{0.61} (.70) &0.94 (.67) & 1.5 (.60) &1.7 (.35) &1.7 (.36)  &1.9 (.37) \\
\cellcolor{cblue!50}$CLC_+$                  &\textnormal{\textit{w}F1}         &1K \textit{im} &89.4 &\textbf{89.5} &85.9  &71.1  &74.6 &73.8 \\
\cellcolor{cblue!50}$PASTIS$                 &\textnormal{\textit{w}F1}          &300K \textit{im} &75.3 &\textbf{80.2} &71.0  &52.2   &56.4  & 54.8 \\
\cellcolor{cblue!50}$PF$                     &\textnormal{\textit{w}F1}         &10K \textit{l}&80.1 &80.0 &\textbf{80.2}  &79.8  &78.5 &78.7 \\
\cellcolor{cgreen!50}$\mathcal{W}^{**}$ & \textnormal{r} & $\approx$3.8M \textit{l} & .75 &--- &--- &--- &--- &--- \\
\bottomrule
\end{tabular}%
\end{sc}
\end{small}
\end{center}
\vskip-0.3cm
\end{table}

\section{Ablation Studies}\label{appendix:ablations}
\subsection{Loss Choice}
We used two different loss functions for the Pixel-Pixel alignment and Pixel-Waveform alignment. \cref{table:cosine_sim_loss} shows that the cosine similarity (CS) using the VICReg loss for the pixel-waveform alignment only reaches a maximum of 0.56  for the positive pairs and 0.10 for the negative pairs, indicating poor alignment. On the other hand, using the ZERO-CL loss, the CS between the positive pairs reaches 0.86, and 0.35 between the negative pairs. We attribute the poor results for Pixel-Waveform alignment using the VICReg loss to the low number of available waveforms within a mini-batch. In contrast, for the Pixel-Pixel alignment, the CS for the positive pairs is 0.99, and 0.04 for the negative pairs using the VICReg loss. It decreases to 0.98 for the positive pairs and 0.45 for the negative pairs with ZERO-CL. Nevertheless, using ZERO-CL for both modalities increased training times exponentially due to the whitening transformation and the high number of available pixels within the mini-batch.         
\begin{table}[h]
\caption{Cosine similarity between positive pairs (+) and negative pairs (-) during pre-training. P represents a pixel embedding, W represents a waveform embedding. Positive Pairs consist of a pixel embedding and its corresponding pixel/waveform embedding, while negative pairs are computed between a pixel embedding and other pixel/waveform embeddings within the mini-batch.\\}
\label{table:cosine_sim_loss}
\centering
\begin{tabular}{lccccc}
\toprule
Loss & +P$\leftrightarrow$P & -P$\leftrightarrow$P & +P$\leftrightarrow$W & -P$\leftrightarrow$W \\
\midrule
VICReg & 0.99 & 0.04 & 0.56 & 0.10 \\
ZERO-CL   & 0.98 & 0.45 & 0.86 & 0.35\\
\bottomrule
\end{tabular}

\end{table}

\subsection{Using a Shared Decoder}
In our formulation, we used two different decoders for horizontal and vertical structural understanding. We hypothesized that a single real-valued embedding cannot simultaneously encode contrasting information, such as trees of the same species with different vertical structures. \cref{table:dual_vs_single_decoder} Shows that in the retrieval case, training a single decoder produces pixel embeddings with no semantic overlap with the waveforms with an average CS of -0.42, while the retrieved pixel embeddings given a pixel input are still highly similar. In the case of training separate decoders, retrieval performance is drastically different, with high correlations (0.99) between pixels and the retrieved waveforms.       
\vspace{-1em}
\begin{table}[h]
\caption{Average cosine similarity ($\overline{CS}$) between retrieved pixels and pixel queries (P$\leftarrow$P) and between retrieved waveforms and pixel queries (P$\leftarrow$W) for KNN=1.\\ }
\label{table:dual_vs_single_decoder}
\centering
\begin{tabular}{lccccc}
\toprule
 & P$\leftarrow$P $\overline{CS}$ & P$\leftarrow$W $\overline{CS}$\\
\midrule
Shared decoder & 0.97 & -0.42\\
Separate decoders   & 0.98 & 0.99\\
\bottomrule
\end{tabular}
\end{table}

\subsection{Hierarchical VICReg Loss}
The original VICReg loss proposed by Bardes et al. \yrcite{bardes2021vicreg} was applied between two views of the same input image. They later extended their work and applied it to local image features \cite{bardes2022vicregl}. Their results showed that better results were obtained when applying the VICReg loss on local features and at the instance level. Here, we further extend this work and apply it hierarchically to the decoder outputs between the pre-trained model and the multi-temporal AE. Results in \cref{table:zero_shot_diff_vicreg} show significant improvements for both datasets when using this formulation. These results are also in line with the findings in Bardes et al. \yrcite{bardes2022vicregl}.   
\vspace{-1em}
\begin{table}[h]
\caption{Zero-shot classification performance ($w$F1) on the $PF$ and $CLC_+$ datasets using a model pre-trained with multiple VICReg losses at different decoder levels ($\text{VICReg}_h$) against a single loss applied on the embeddings from the last decoder layer ($\text{VICReg}_s$).}
\vspace{0.5em}
\label{table:zero_shot_diff_vicreg}
\centering
\begin{tabular}{lccccc}
\toprule
 & $PF$ & $CLC_+$\\
\midrule
$\text{VICReg}_h$   & 76.0 & 80.1\\
$\text{VICReg}_s$   & 67.6 & 74.2\\
\bottomrule
\end{tabular}

\end{table}
\vspace{-1em}
\subsection{Neighborhood Attention}
The output of our pre-trained model is composed of two neighborhood attention layers per output head, instead of the standard convolutional layers. This design choice allows each embedding to be modeled based on its local neighborhood instead of relying on small and fixed receptive fields. \cref{table:na_perf} shows an increase of 2.1\% ($w$F1) for the $CLC_+$ dataset and an RMSE decrease of 0.4 m for the $\mathcal{W}_{rh}$ dataset using this new configuration.       
\vspace{-1em}
\begin{table}[h]
\caption{Performance differences between a convolutional output layer (CNN) vs. a NA layer on the $CLC_+$ and $\mathcal{W}_{rh}$ datasets.}
\vspace{0.5em}
\label{table:na_perf}
\centering
\begin{tabular}{lccccc}
\toprule
 & $CLC_+$ & $\mathcal{W}_{rh}$\\
\midrule
CNN   & 72.1 (\textnormal{\textit{w}F1}) & 2.4 (RMSE)\\
NA   & 74.2 (\textnormal{\textit{w}F1}) & 2.0 (RMSE)\\
\bottomrule
\end{tabular}
\end{table}

\subsection{Embedding sensitivity}
Our results, either in the zero-shot setting or the fine-tuned setting used embeddings from $\mathcal{O}^{\mathcal{H}}$ for horizontal structure-related products (e.g., tree species identification, land cover mapping) and embeddings from $\mathcal{O}^{\mathcal{V}}$ for vertical structure-related products (e.g., canopy height mapping, waveform retrieval). We perform the same tests in the zero-shot setting, but reversing the query and retrieval embeddings for the other structure direction. \cref{table:reversed_embeddings} shows that relying on vertical structure data for products requiring horizontal understanding and vice versa performs poorly for the six tested products. This validates our design choice of cross-modal alignment with vertical structure data for EO products relying on this type of information.

\begin{table}[H]
\caption{Top-1 retrieval-based zero-shot classification performance of DUNIA with opposite direction query and retrieval embeddings. OG represents the original results from \cref{table:retrieval_performance} for KNN=50. \colorbox{cgreen!50}{\textcolor{cgreen!50}{---}} and \colorbox{cblue!50}{\textcolor{cblue!50}{---}} are DUNIA query embeddings from $\mathcal{O^V}$ and $\mathcal{O^H}$ respectively.}
\label{table:reversed_embeddings}
\begin{center}
\begin{small}
\begin{sc}

\begin{tabular}{lcccccccc}
\toprule
Dataset & Metric & OG & DUNIA                                  \\
\midrule                         
\cellcolor{cblue!50}$\mathcal{W}_{rh}$                 &\textnormal{RMSE (r)}             &2.0 (.93)   &4.3 (.63) \\
\cellcolor{cblue!50}$\mathcal{W}_c$                    &\textnormal{RMSE (r)}            &11.7 (.89)  &18.9 (.53) \\
\cellcolor{cblue!50}$\mathcal{W}_{pai}$                &\textnormal{RMSE (r)}            & 0.71 (.75) &1.6 (.42) \\
\cellcolor{cgreen!50}$CLC_+$                  &\textnormal{\textit{w}F1}           &80.1        &35.1\\
\cellcolor{cgreen!50}$PASTIS$                 &OA    &56.2  &0.0\\
\cellcolor{cgreen!50}$PF$                     &\textnormal{\textit{w}F1}            &73.8        &56.1 \\
\bottomrule
\end{tabular}
\end{sc}
\end{small}
\end{center}
\end{table} 

\subsection{Inference Runtime (Zero-Shot)}
Generating maps in the zero-shot setting requires: 1) a forward pass through the pre-trained model (encoder-dual decoders) to obtain the embeddings, 2) for each pixel, retrieve the $k$ nearest neighbors (KNN) from the retrieval database (DB), and 3) classify the queried pixel embedding based on distance-weighted voting. As such, generating a map over e.g. a $ 20.48 \times 20.48$ Km area at 10 m resolution, requires querying and classifying 4,194,304 pixels. However, \cref{table:wall_clock_times} shows that the querying and classification times only add a small overhead. Moreover, the time required for the three operations (forward pass, retrieval, and classification) is still faster than a similarly capable model like AnySat.      
\vspace{-1em}
\begin{table}[H]
\caption{
Inference runtime (in seconds) of DUNIA in the zero-shot setting, compared to AnySat. 
DB denotes the retrieval database of $N$ key/value pairs (embedding/label), and 
KNN is the number of nearest neighbors retrieved per pixel. 
All times are in seconds.}
\label{table:wall_clock_times}
\begin{center}
\begin{small}
\begin{sc}

\begin{tabular}{lcccccc}
\toprule
Model & DB & KNN & Forward pass & Retrieval & Classification & Total \\
\midrule
DUNIA & 256K & 100 & 2.52 & 0.36 & 1.34 & 4.22 \\
DUNIA & 512K & 200 & 2.52 & 0.40 & 1.88 & 4.80 \\
AnySat & --- & --- & 177.37 & --- & --- & 177.37 \\
\bottomrule
\end{tabular}
\end{sc}
\end{small}
\end{center}
\end{table}

\subsection{Impact of Input Size}

To assess the sensitivity of DUNIA to the input image resolution, we conducted experiments using three different input sizes: $128 \times 128$, $256 \times 256$, and $512 \times 512$. As shown in Table~\ref{table:effect_input_resolution}, the results across datasets and metrics remain unchanged. This confirms that image size has a negligible impact on the final performance of the model in the zero-shot setting.
\vspace{-1em}
\begin{table}[H]
\caption{Zero-shot performance of DUNIA at different input image sizes for several datasets.}
\vspace{0.5em}
\label{table:effect_input_resolution}
\begin{center}
\begin{small}
\begin{sc}
\begin{tabular}{lccccc}
\toprule
Dataset & Metric & 128$\times$128 & 256$\times$256 & 512$\times$512 \\
\midrule
$\mathcal{W}_{rh}$ & \textnormal{RMSE (r)}   & 2.2   & 2.0   & 2.1 \\
$\mathcal{W}_{c}$   & \textnormal{RMSE (r)}   & 11.6  & 11.7  & 11.7 \\
$CLC+$    & \textnormal{\textit{w}F1}    & 80.2 & 80.1 & 80.2 \\
\bottomrule
\end{tabular}
\end{sc}
\end{small}
\end{center}
\end{table}

\subsection{Temporal Stability}

To evaluate the temporal stability of DUNIA, we conducted two experiments on several vertical structural products across the years 2019, 2020, and 2021. Other products were excluded due to the unavailability of labels for different years.

1. Fine-tuning an MLP head on top of a model pre-trained in 2020, and evaluating on 2019 and 2021.\\
2. Pre-training DUNIA using combined data from 2019–2021 and evaluating it in the zero-shot setting.

Table~\ref{table:temporal_finetune} shows that fine-tuning on individual years yields consistent performance across time. Similarly, Table~\ref{table:temporal_zeroshot} demonstrates that zero-shot performance is also stable over several years.

\begin{table}[H]
\caption{Fine-tuned performance (RMSE) on vertical structure variables estimation across several years.}
\vspace{0.5em}
\label{table:temporal_finetune}
\begin{center}
\begin{small}
\begin{sc}
\begin{tabular}{lcccc}
\toprule
Dataset & Metric & 2019 & 2020 & 2021 \\
\midrule
$\mathcal{W}_{rh}$  & RMSE   & 1.35 & 1.34 & 1.40 \\
$\mathcal{W}_{c}$  & RMSE   & 9.8 & 9.8 & 10.1 \\
$\mathcal{W}_{pai}$  & RMSE   & 0.65 & 0.62 & 0.63 \\
\bottomrule
\end{tabular}
\end{sc}
\end{small}
\end{center}
\end{table}

\vspace{-2em}

\begin{table}[H]
\caption{Zero-shot performance (RMSE) on vertical structure variables retrieval across years.}
\vspace{0.5em}
\label{table:temporal_zeroshot}
\begin{center}
\begin{small}
\begin{sc}
\begin{tabular}{lcccc}
\toprule
Dataset & Metric & 2019 & 2020 & 2021 \\
\midrule
$\mathcal{W}_{rh}$  & RMSE   & 2.4 & 2.0 & 2.1 \\
$\mathcal{W}_{c}$  & RMSE   & 11.6 & 11.7 & 11.9 \\
$\mathcal{W}_{pai}$  & RMSE   & 0.75 & 0.71 & 0.74 \\
\bottomrule
\end{tabular}
\end{sc}
\end{small}
\end{center}
\end{table}

\subsection{Impact of S-2 and S-2 Input Images}

\cref{table:optical_radar_finetune} shows that combining both modalities yields the best performance for height estimation and tree species identification, while land cover classification is relatively robust to the choice of modality.

\vspace{-1em}
\begin{table}[H]
\caption{Performance of DUNIA when using S-1 only, S-2 only, or both modalities. Results are reported for height prediction (RMSE), land cover classification ($w$F1) on the $CLC_+$ dataset, and tree species identification ($w$F1) on the $PF$ dataset. Performance results obtained in the fine-tuned setting.}
\vspace{0.5em}
\label{table:optical_radar_finetune}
\begin{center}
\begin{small}
\begin{sc}
\begin{tabular}{lcccc}
\toprule
Dataset & Metric & S-1 Only & S-2 Only & S-2 \& 1 \\
\midrule
$\mathcal{W}_{rh}$  & RMSE   & 3.80 & 2.80 & 1.34 \\
$PF$  & \textnormal{\textit{w}F1}   & 65.5\% & 81.2\% & 82.2\% \\
$CLC_+$  & \textnormal{\textit{w}F1}   & 90.2\% & 90.0\% & 90.3\% \\
\bottomrule
\end{tabular}
\end{sc}
\end{small}
\end{center}
\end{table}

\subsection{Median Composites vs. Single-Date Images}
Results on height estimation and crop classification ($PASTIS$) (\cref{table:median_vs_single_images}) show that median composites — even without full time series — significantly outperform single-date imagery by capturing more stable and representative reflectance patterns over time.
\vspace{-1em}
\begin{table}[H]
\caption{Comparison between single-date and median composite inputs for products where temporal information is critical. Performance results obtained in the fine-tuned setting.}
\vspace{0.5em}
\label{table:median_vs_single_images}
\begin{center}
\begin{small}
\begin{sc}
\begin{tabular}{lccc}
\toprule
Dataset & Metric & Single-date image & Median composite \\
\midrule
$\mathcal{W}_{rh}$  & $\text{RMSE}$   & 1.9 & 1.3 \\
$PASTIS$  & \textnormal{\textit{w}F1}   & 42.3  & 77.0 \\
\bottomrule
\end{tabular}
\end{sc}
\end{small}
\end{center}
\end{table}
\clearpage
\section{Further Results}
\begin{figure}[!h]
\begin{center}
\resizebox{0.9\textwidth}{!}{%
\includegraphics{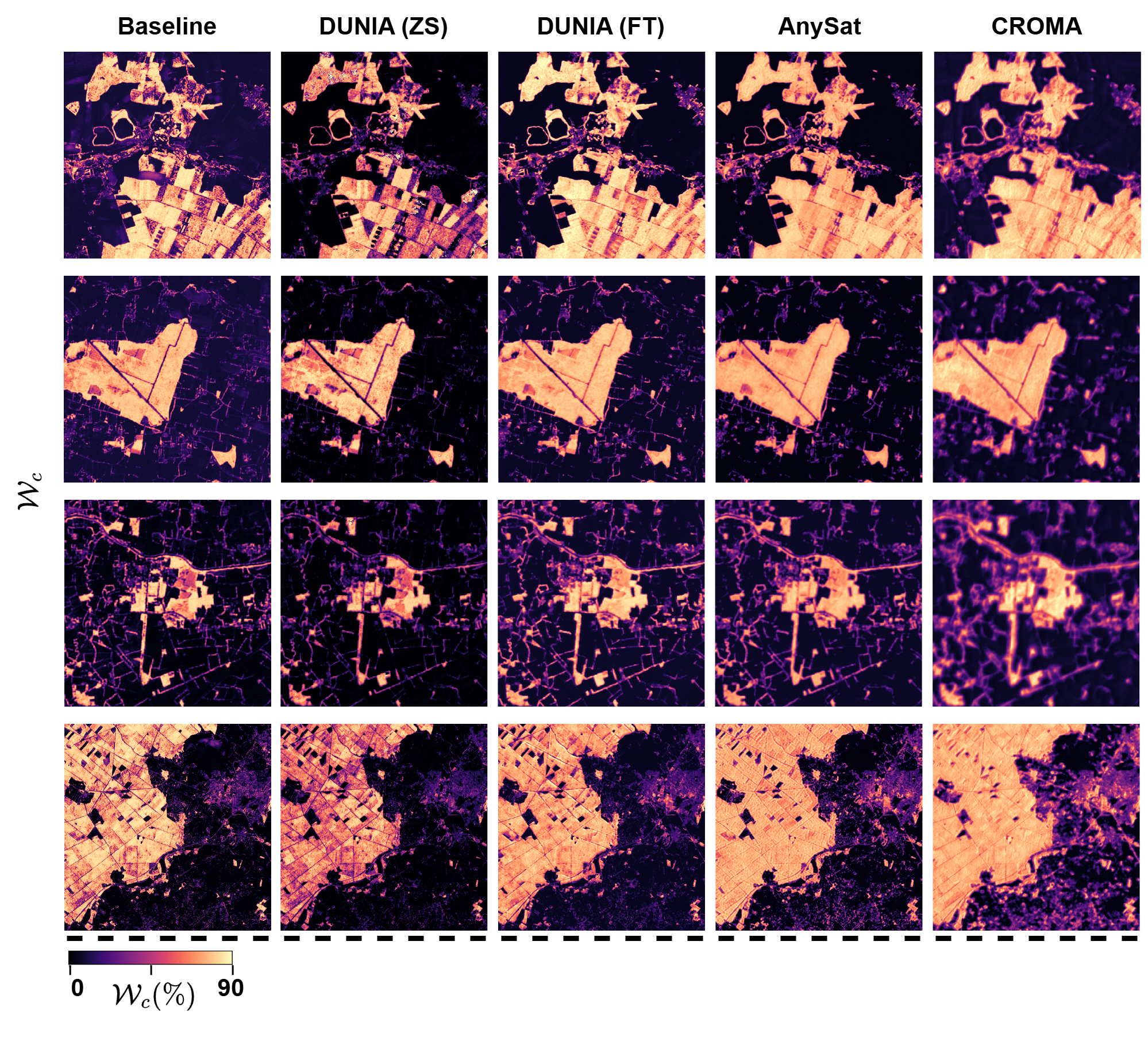}%
}
\end{center}
\vspace{-3em}
\caption{Fractional canopy cover ($\mathcal{W}_{c}$) maps produced with different models. Baseline represents reference maps from adapted FORMS \cite{schwartz2023forms}. DUNIA (ZS) represents maps obtained in the zero-shot setting, while DUNIA (FT) represents maps obtained in the fined-tuned setting. Best viewed zoomed-in (300+\%).}
\label{fig:result_maps_canopy_cover}
\end{figure}

\clearpage
\begin{figure*}[h!t]
\begin{center}
\resizebox{0.9\textwidth}{!}{%
\includegraphics{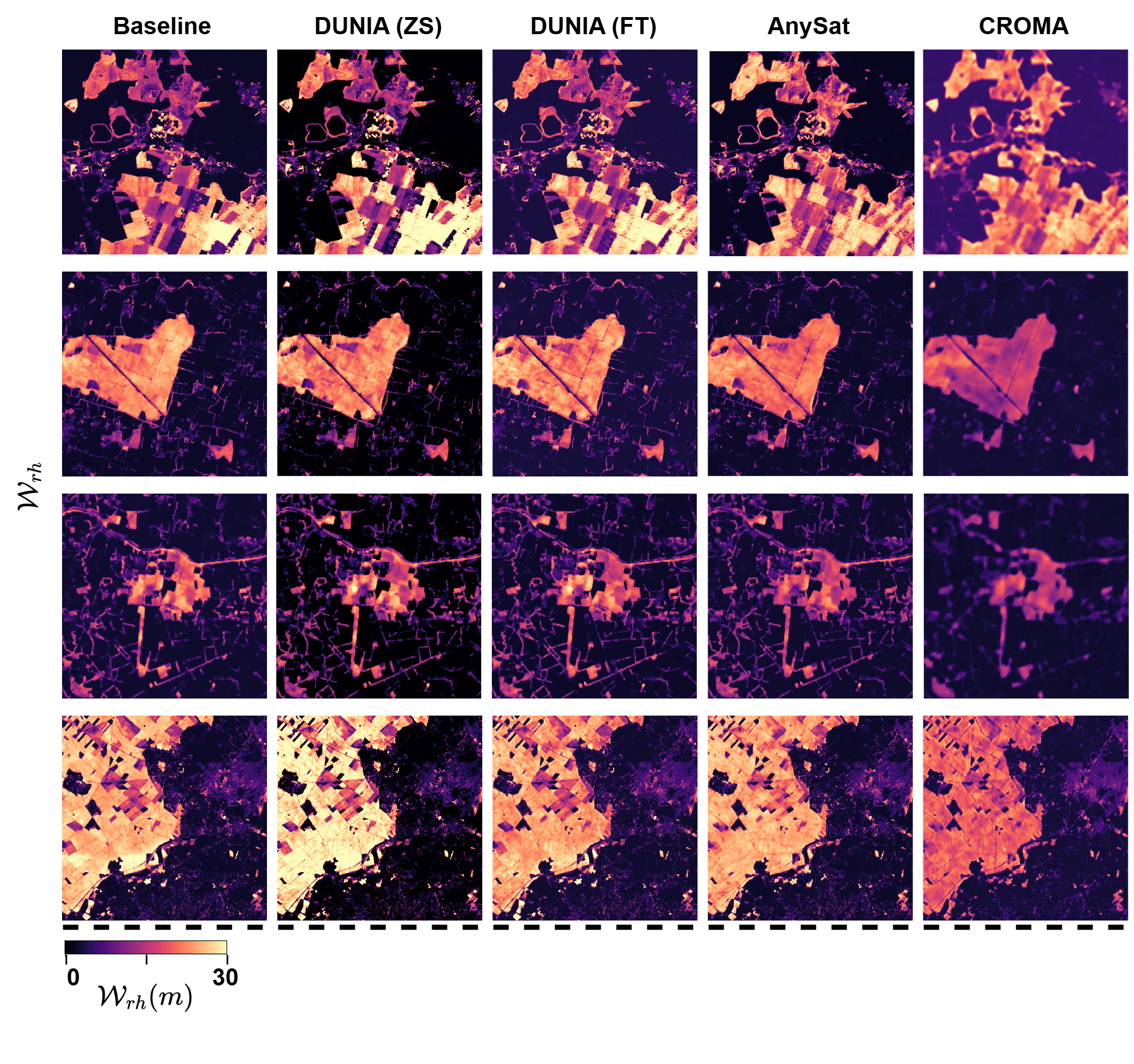}%
}
\end{center}
\vspace{-3em}
\caption{Canopy height ($\mathcal{W}_{rh}$) maps produced with different models. Baseline represents reference maps from FORMS \cite{schwartz2023forms}. DUNIA (ZS) represents maps obtained in the zero-shot setting, while DUNIA (FT) represents maps obtained in the fined-tuned setting. Best viewed zoomed-in (300+\%).}
\label{fig:result_maps_canopy_height}
\end{figure*}

\clearpage
\begin{figure}[h!t]
\begin{center}
\resizebox{0.9\textwidth}{!}{%
\includegraphics{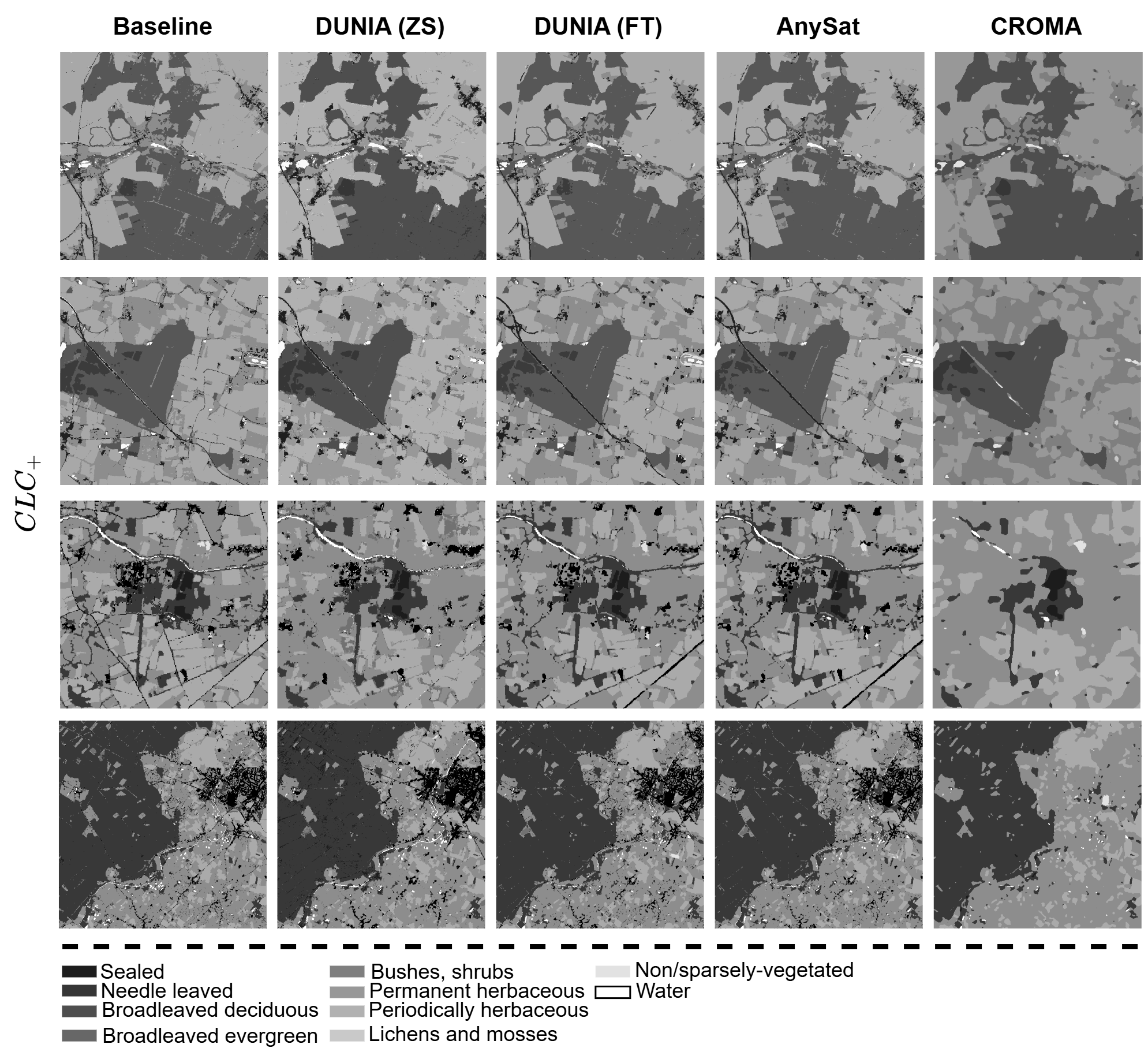}%
}
\end{center}
\vspace{-1em}
\caption{Land cover classes ($CLC_+$) maps produced with different models. Baseline represents reference maps from the \citet{clc}. DUNIA (ZS) represents maps obtained in the zero-shot setting, while DUNIA (FT) represents maps obtained in the fined-tuned setting. Best viewed zoomed-in (300+\%).}
\label{fig:result_maps_clc}
\end{figure}

\clearpage
\begin{figure}[h!t]
\begin{center}
\resizebox{\textwidth}{!}{%
\includegraphics{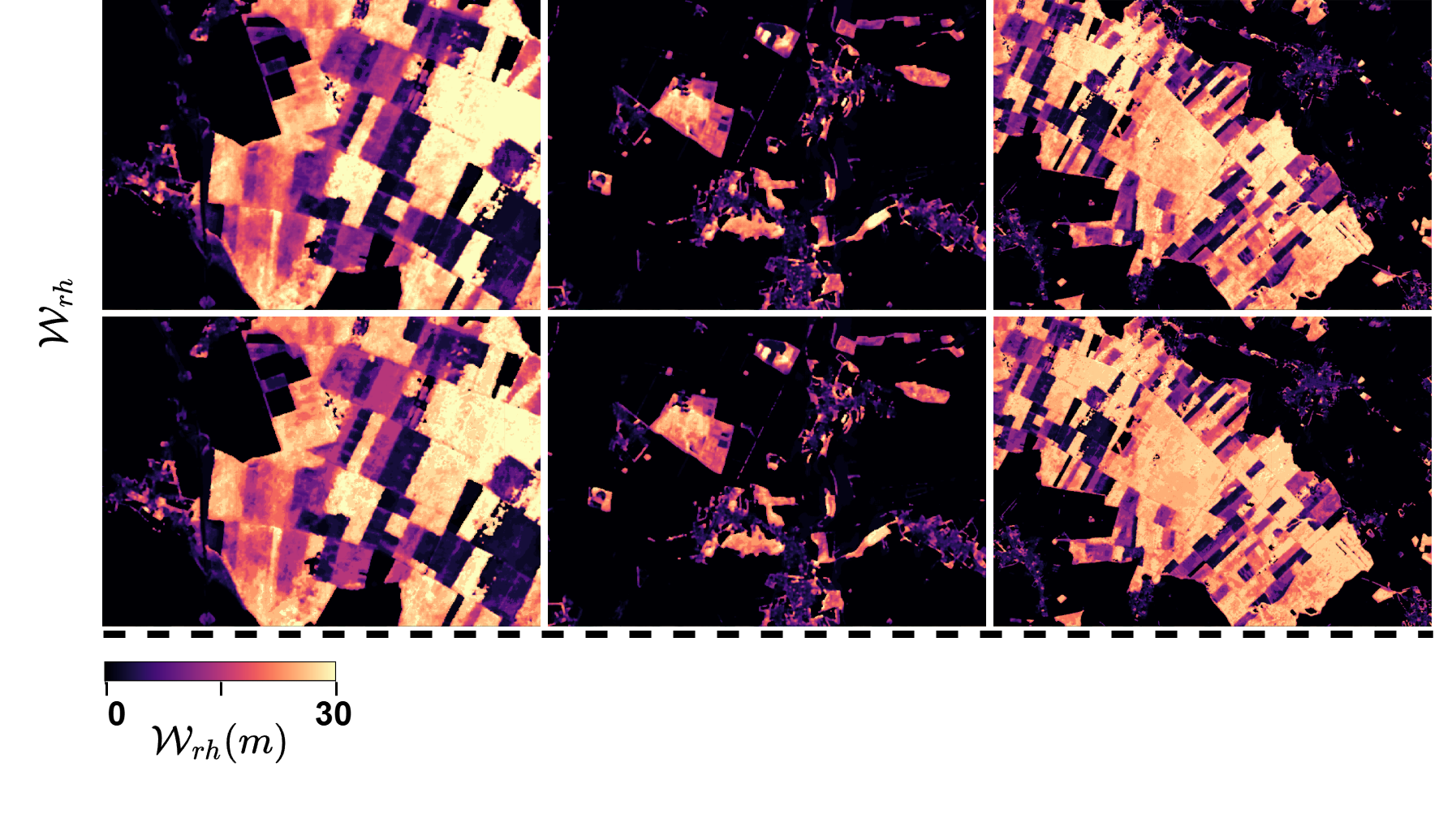}%
}
\end{center}
\vspace{-4em}
\caption{Comparison of map qualities for the task of estimating the canopy height in the zero-shot setting using 50K $l$ GEDI samples as queries (top row) vs. only 10K $l$ (bottom row).}
\label{fig:high_vs_low_samples}
\end{figure}

\definecolor{pyblue}{RGB}{31, 119, 180}
\definecolor{pygreen}{RGB}{44, 160, 44}
\definecolor{pygrey}{RGB}{128, 128, 128}

\newcommand{\sinewavelegend}[2][pyblue]{
    \tikz[baseline=-0.2ex, scale=#2]{
        \draw[thick, #1, samples=50, domain=0:pi] plot (\x,{sin(\x r)});
        \draw[thick, #1, samples=50, domain=0:pi, xshift=3cm, yshift=0.15cm, xscale=0.5, yscale=1.5] plot (\x,{sin(\x r)});
    }
}

\begin{figure*}[ht]
\begin{center}
\resizebox{0.90\textwidth}{!}{%
\includegraphics{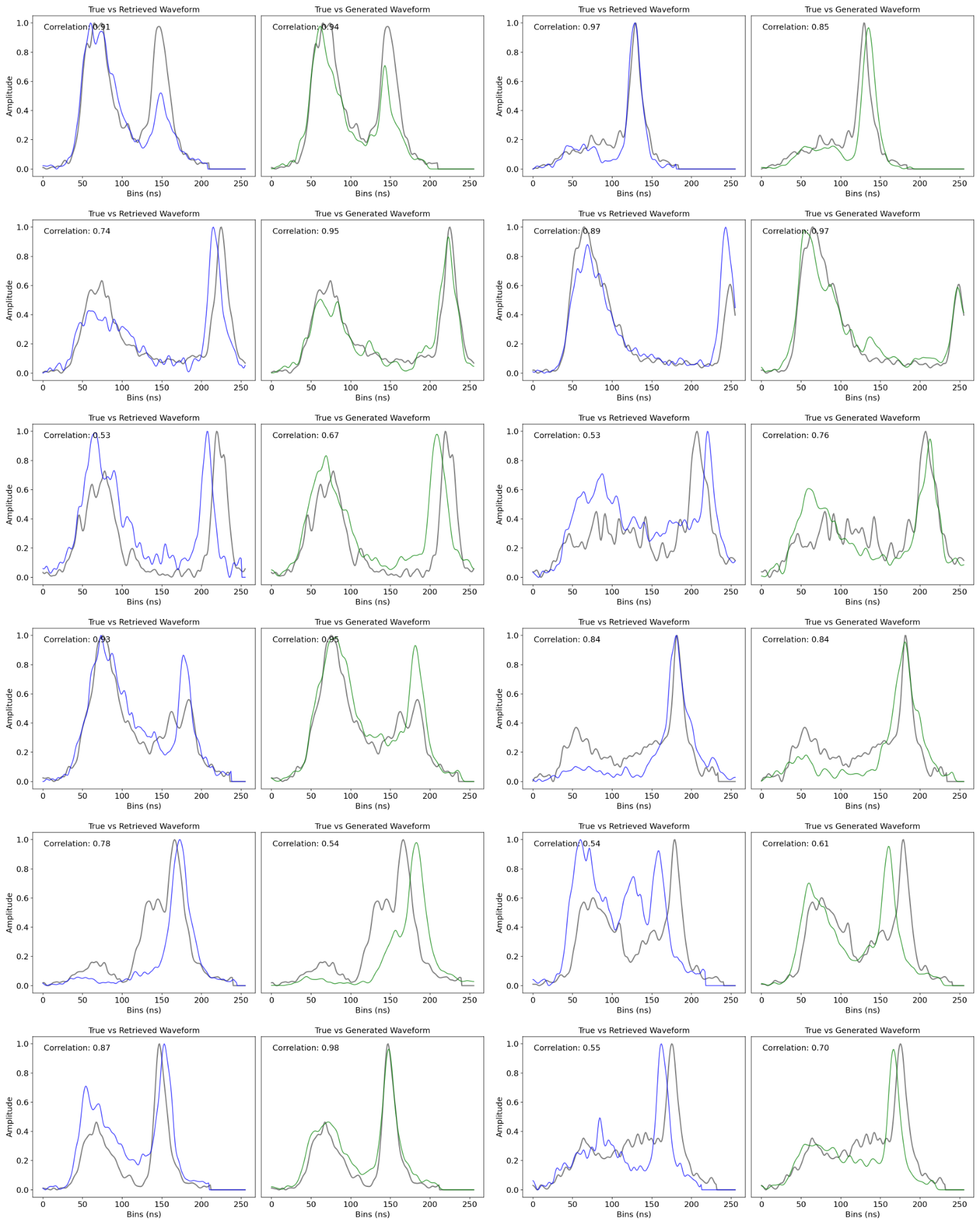}%
}
\end{center}
\caption{Uncurated list of retrieved (\sinewavelegend[pyblue]{0.1}) and generated waveforms (\sinewavelegend[pygreen]{0.1}) overlayed on a reference waveform (\sinewavelegend[pygrey]{0.1}). 'Correlation' is Pearson's correlation coefficient (r) between the reference and the retrieved/generated waveforms. Best viewed zoomed-in (200+\%).}
\label{fig:result_waveforms}
\end{figure*}

\begin{figure*}[ht]
\begin{center}
\resizebox{0.90\textwidth}{!}{%
\includegraphics{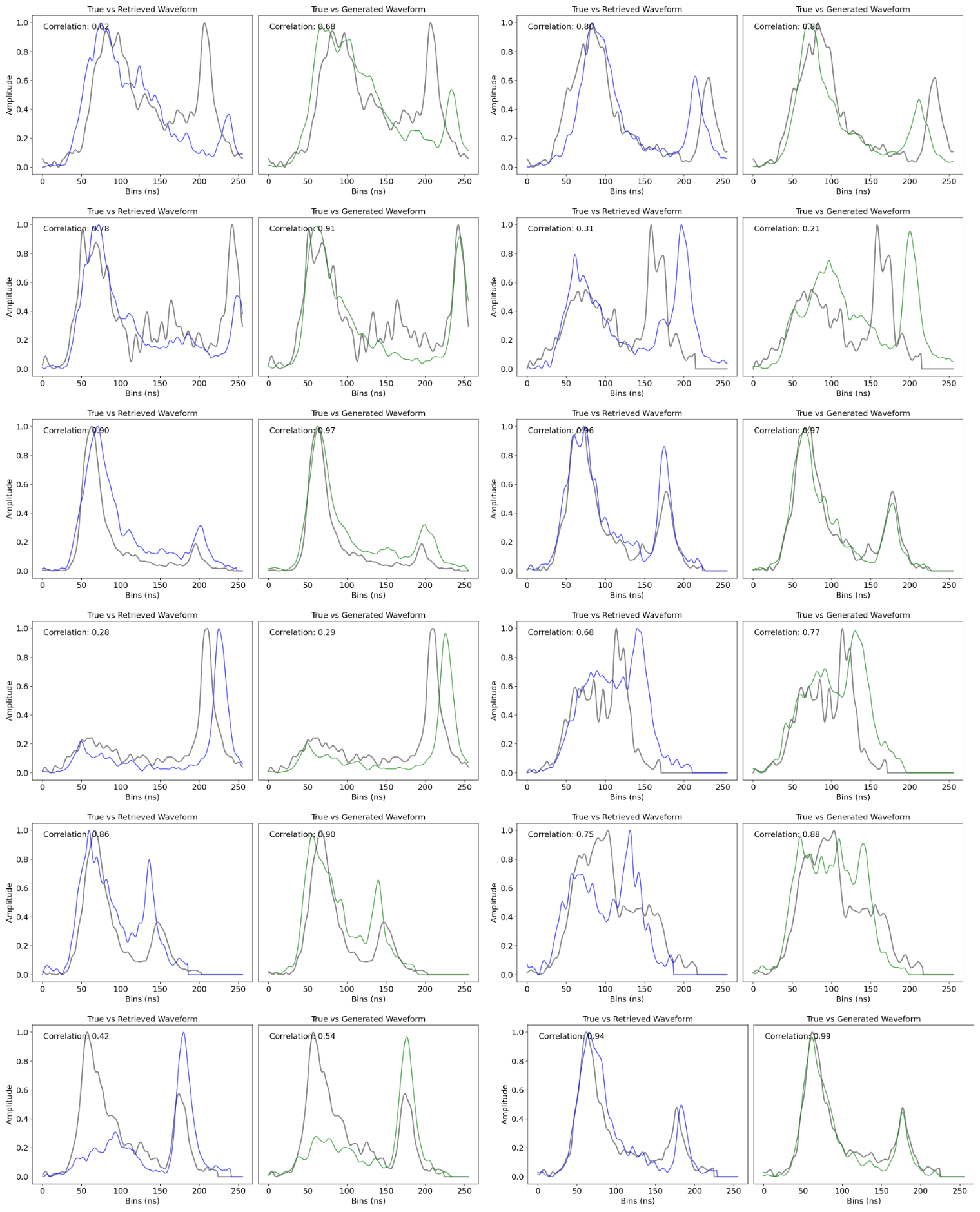}%
}
\end{center}
\caption{Uncurated list of retrieved (\sinewavelegend[pyblue]{0.1}) and generated waveforms (\sinewavelegend[pygreen]{0.1}) overlayed on a reference waveform (\sinewavelegend[pygrey]{0.1}). 'Correlation' is Pearson's correlation coefficient (r) between the reference and the retrieved/generated waveforms. Best viewed zoomed-in (200+\%).}
\label{fig:result_waveforms_set_2}
\end{figure*}

\begin{figure*}[ht]
\begin{center}
\resizebox{0.90\textwidth}{!}{%
\includegraphics{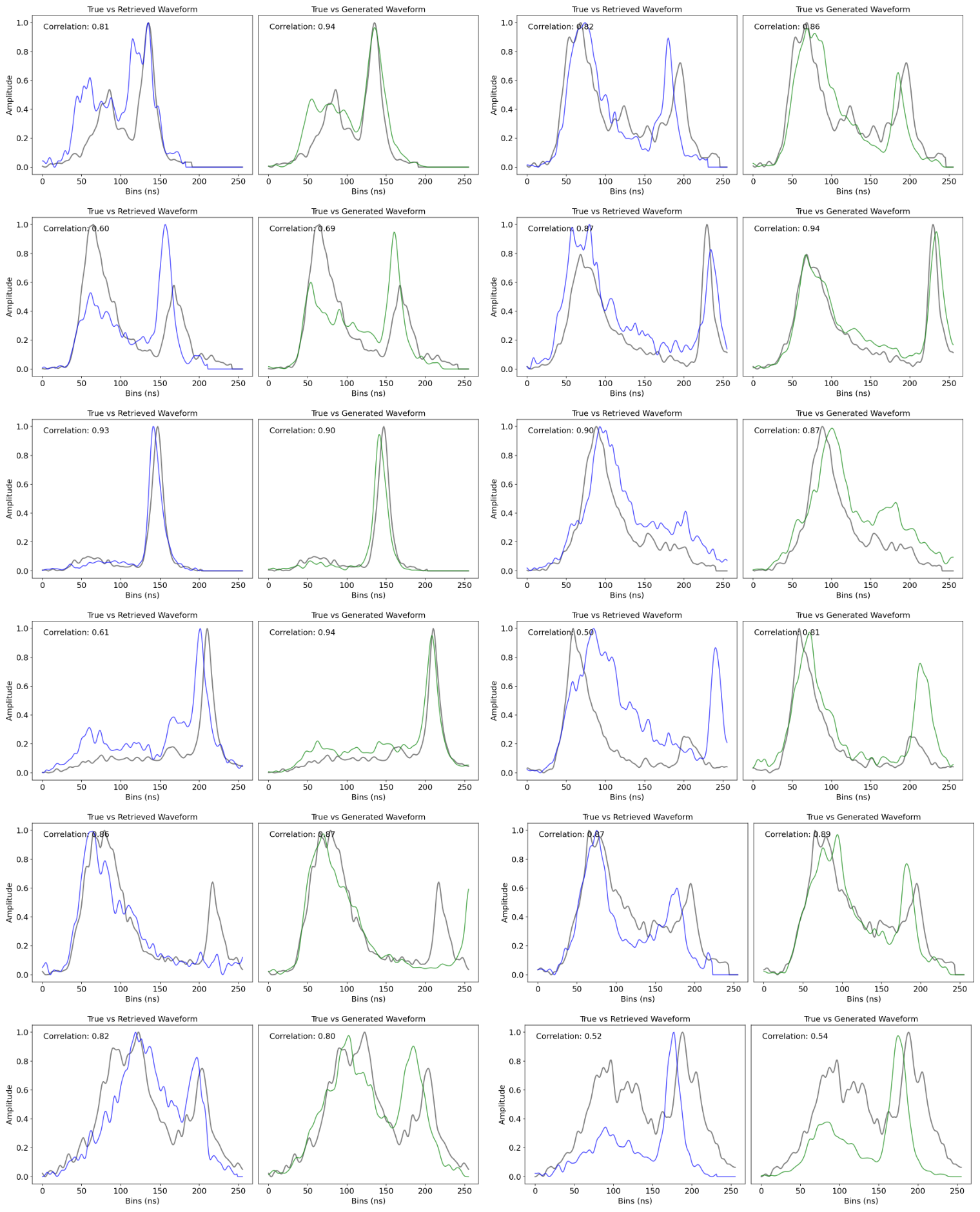}%
}
\end{center}
\caption{Uncurated list of retrieved (\sinewavelegend[pyblue]{0.1}) and generated waveforms (\sinewavelegend[pygreen]{0.1}) overlayed on a reference waveform (\sinewavelegend[pygrey]{0.1}). 'Correlation' is Pearson's correlation coefficient (r) between the reference and the retrieved/generated waveforms. Best viewed zoomed-in (200+\%).}
\label{fig:result_waveforms_set_3}
\end{figure*}

\begin{figure*}[ht]
\begin{center}
\resizebox{0.90\textwidth}{!}{%
\includegraphics{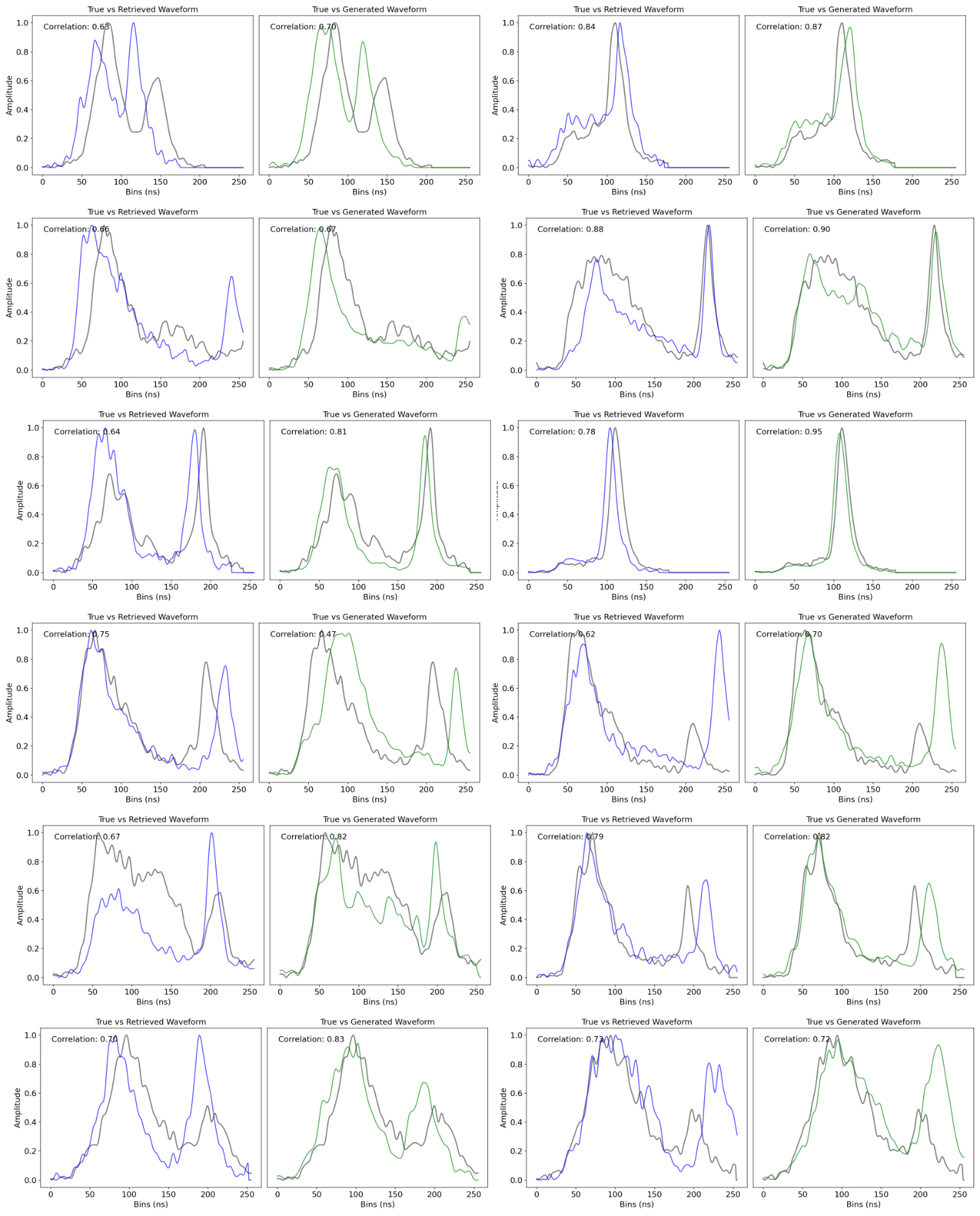}%
}
\end{center}
\caption{Uncurated list of retrieved (\sinewavelegend[pyblue]{0.1}) and generated waveforms (\sinewavelegend[pygreen]{0.1}) overlayed on a reference waveform (\sinewavelegend[pygrey]{0.1}). 'Correlation' is Pearson's correlation coefficient (r) between the reference and the retrieved/generated waveforms. Best viewed zoomed-in (200+\%).}
\label{fig:result_waveforms_set_4}
\end{figure*}

\end{document}